\documentclass[journal]{IEEEtran}

\usepackage{graphicx}
\usepackage{amsmath}
\usepackage{amsfonts}
\usepackage{multirow}
\usepackage{color}
\usepackage{soul}
\usepackage{algorithm} 
\usepackage{algorithmic} 
\usepackage{xcolor}
\usepackage{slashbox}
\usepackage{stfloats}
\usepackage{subfigure}
\usepackage{cite}

\begin{document}

\title{SCE: A Manifold Regularized Set-Covering Method for Data Partitioning}

\author{Xuelong Li,~\IEEEmembership{Fellow,~IEEE}, Quanmao Lu, Yongsheng Dong,~\IEEEmembership{Member,~IEEE}, and Dacheng Tao,~\IEEEmembership{Fellow,~IEEE} 
\thanks{This work was supported in part by the National Natural
Science Foundation of China under Grant U1604153 and 61125106,
in part by the International Science and Technology Cooperation Project
of Henan Province under Grant 162102410021, in part by the State Key Laboratory of
Virtual Reality Technology and Systems under Grant BUAA-VR-16KF-04, in part by the Key Laboratory of Optoelectronic Devices and Systems
of Ministry of Education and Guangdong Province under Grant GD201605, and in part by Australian Research Council Projects
FT-130101457, DP-140102164 and  LP-150100671. (Corresponding author: Yongsheng Dong.)

X. Li is with the Center for OPTical IMagery Analysis and Learning (OPTIMAL), State Key Laboratory of Transient
Optics and Photonics, Xi'an Institute of Optics and Precision Mechanics, Chinese Academy of Sciences, Xi'an 710119, China (email: xuelong\_li@opt.ac.cn).

Q. Lu is with the Center for OPTical IMagery Analysis and Learning (OPTIMAL), State Key Laboratory of Transient
Optics and Photonics, Xi'an Institute of Optics and Precision Mechanics, Chinese Academy of Sciences, Xi'an 710119, China, and also with University of Chinese Academy of Sciences, Beijing 100049, China (email: quanmao.lu.opt@gmail.com).

Y. Dong is with the Center for Optical Imagery Analysis and Learning, State Key Laboratory of Transient Optics and Photonics, Xi¡¯an Institute of Optics and Precision Mechanics, Chinese Academy of Sciences, Xi¡¯an 710119, China, and also with the Information Engineering College, Henan University of Science and Technology, Luoyang 471023, China (e-mail: dongyongsheng98@163.com).

D. Tao is with the UBTech Sydney Artificial Intelligence Institute and the School of Information Technologies in the Faculty of Engineering and Information Technologies at The University of Sydney, J12/318 Cleveland St, Darlington NSW 2008, Australia (email: dacheng.tao@sydney.edu.au).

\copyright 20XX IEEE. Personal use of this material is permitted. Permission from IEEE must be obtained for all other uses, in any current or future media, including
reprinting/republishing this material for advertising or promotional purposes, creating new collective works, for resale or redistribution to servers or lists, or reuse of any copyrighted component of this work in other works.

}
}

\markboth{Transactions on Neural Networks and Learning Systems}%
{Shell \MakeLowercase{\textit{et al.}}: Bare Demo of IEEEtran.cls for Journals}
\maketitle

\begin{abstract}
Cluster analysis plays a very important role in data analysis. In these years, cluster ensemble, as a cluster analysis tool, has drawn much attention for its robustness, stability and accuracy. Many efforts have been done to combine different initial clustering results into a single clustering solution with better performance.
However, they neglect the structure information of the raw data in performing the cluster ensemble. In this paper, we propose a  Structural Cluster Ensemble (SCE) algorithm for data partitioning formulated as a set-covering problem. In particular, we construct a Laplacian regularized objective function  to capture the structure information among clusters. Moreover, considering the importance of the discriminative information underlying in the initial clustering results, we add a discriminative constraint into our proposed objective function.
Finally, we verify the performance of the SCE algorithm on both synthetic and real data sets. The experimental results show the effectiveness of our proposed SCE algorithm.
\end{abstract}

\begin{IEEEkeywords}
Cluster ensemble, set-covering, manifold structure, discriminative constraint.
\end{IEEEkeywords}

\IEEEpeerreviewmaketitle

\section{Introduction}

\IEEEPARstart{D}{ata} clustering is an essential technique in many fields, such as data mining \cite{huang2014extensions},
pattern recognition \cite{duda1973pattern}, image segmentation \cite{frigui1999robust}, image retrieval \cite{philbin2008lost} and data compression \cite{gersho1992vector}. It aims to find the underlying structure of a given dataset based on various criteria, obtaining "the proper" clustering result. Compared with supervised learning \cite{7159100, 7172530, 7437460, 7398054}, clustering analysis always has no prior information which increases its difficulty to divide data into several clusters.

During these decades, lots of clustering algorithms \cite{jain1999data,bassani2015dimension,hou2014discriminative,elkan2003using,jing2013dictionary,LiuTX16} have been proposed to solve this hard problem, including K-means, Gaussian Mixture Model (GMM), spectral clustering \cite{von2007tutorial} and hierarchical clustering algorithm, etc. However, for the same dataset, different clustering methods or even the same method with different parameters usually produce different clustering results (or partitions). Single-run clustering algorithms are unstable and heavily depend on the dataset.
Therefore, we can not obtain the satisfactory clustering results of all types of datasets only by using a single clustering algorithm.

\begin{figure}[h]
\begin{center}
\includegraphics[width=3.2in]{{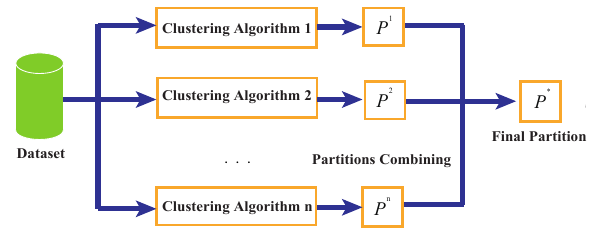}}
\end{center}
\caption{The flowchart of cluster ensemble.}
\label{li1}
\end{figure}

There are many feasible approaches to improve the performance of clustering analysis. Among them, cluster ensemble is an effective and mostly used approach. Particularly, cluster ensemble combines the multiple clustering solutions into a single consensus one, and has two main steps: obtaining initial partitions with all kinds of clustering algorithms and integrating all partitions into a new partition which is the final result. Fig. \ref{li1} gives the whole framework of cluster ensemble \cite{ghaemi2009survey}. In the first step, different clustering algorithms are used to obtain the initial partitions of the raw data. In the second step, a heuristic method \cite{strehl2002cluster} or consensus function \cite{mirkin2001reinterpreting,topchy2004mixture,wang2011bayesian} can be used to obtain a final partition with higher consistency and stability.

Despite the low performance of weak clustering algorithms,  Topchy et al. showed that an effective and stable partition can be produced if a proper consensus function can be used to perform cluster ensemble \cite{topchy2003combining, fern2003random}. Kuncheva and Hadjitodorov \cite{kuncheva2004using} discussed the importance of diversity in generation step and proposed a variant of the generic ensemble method where the number of clusters is chosen randomly for every initial partition. From its experimental results, we can see its capability of achieving a better match with the known cluster label. Inspired by these works, we in this paper use the Max-Min Distance Algorithm (MMDA) \cite{Jia2010Clustering}
to obtain different initial clustering results. MMDA can be regarded as a weak clustering method and has low time complexity. Furthermore, we choose different number of clusters closing to the final number of clusters for increasing the diversity in partitions when producing every ensemble member.

In order to make use of the information provided by different initial partitions, many ensemble methods were proposed to obtain consensus clustering, including relabeling and voting \cite{dudoit2003bagging}, constructing co-association matrix \cite{fred2001finding}, employing graph and hypergraph \cite{domeniconi2009weighted} and other methods \cite{vega2011survey}. However, they neglect the structure information of the raw data in performing the cluster ensemble. Motivated by this problem, we in this paper propose a Structural Cluster Ensemble (SCE) algorithm to perform data partitioning. Particularly, we first formulate the cluster ensemble problem as a set-covering one.
For making the most use of the information provided by the initial partitions and obtaining high quality fusion result, a Laplacian matrix is constructed by using the similarity between initial clusters. Furthermore, the corresponding Laplacian regularization is added to the objective function for capturing the manifold structure lying in the raw data \cite{cai2011graph}. Besides, we take the discriminative information among the raw data into account. It can guarantee the clusters we chosen have high dispersion degree.
In our experiments, we use the Minimum Sum-of-Squares Clustering (MSSC) criterion to evaluate the performance of our method. Experimental results reveal the effectiveness of our proposed. The contributions of this paper are as follows:
\begin{enumerate}
\item We propose a structure cluster ensemble method to partition the raw data. This method can capture the structure information of the raw data by optimizing a manifold regularized objective function.
\item  We use the  discriminative information among the raw data in performing cluster ensemble. It is implemented by constraining the inter-class distance of the initial clusters in our proposed objective function.
Many cluster ensemble methods overlook the discriminative information of the raw data.
\item Our proposed structure cluster ensemble method is an effective and robust one. Experimental results on synthetically generated data and real data sets demonstrate the effectiveness of our proposed SCE method.
\end{enumerate}

The rest of this paper is arranged as below: Section 2 gives the formal definition of cluster ensemble and reviews several representative cluster ensemble methods.
Section 3 describes the set partition problem and our proposed SCE algorithm. The experimental results on both synthetic and real datasets are presented in Section 4.
The paper is briefly concluded in Section 5.

%

\section{Related work}
Cluster ensemble is an important method for improving clustering performance. Recently, many cluster ensemble algorithms have been proposed and applied for many areas including image segmentation \cite{hong2008unsupervised}, data mining \cite{jain1999data} and so on \cite{fischer2003bagging,azimi2006clustering}. Before reviewing the related work, we first give a brief description of cluster ensemble.

Let $X=\{x_1,x_2,...,x_n\}$ represents a set of $n$ objects in a $d$-dimensional space. A clustering algorithm regards $X$ or part of $X$ as input and outputs $k$ clusters about $n$ objects in $X$. To perform cluster ensemble, multiple clustering results need be obtained by implementing a single clustering algorithm several times or multiple clustering algorithms on the raw data. Let $\mathbf{P}=\{P^1,P^2,...,P^M\}$ denotes the $M$ initial clustering results.
Each component partition in $\mathbf{P}$ is a set of non-overlapping and finite clusters with $P^i=\{C_1^i,C_2^i,...,C_{k_i}^i\}$, $X=C_1^i\bigcup{...}\bigcup{C_{k_i}^i}$, and $k_i$ is the number of clusters in the $i$-th partition. We assume that $P^j(x_i)$ denotes a label assigned to $x_i$ in the $j$-th partition and $c_i^j$ represents the center of the cluster $C_i^j$.
The purpose of cluster ensemble is to find a more robust and effective partition $P^*$ of the data $X$ that summarizes the information from partitions in $\mathbf{P}$, where $P^*=\{C_1^*,C_2^*,...,C_{k^*}^*\}$ and $k^*$ is the final number of clusters. Next, we review some recent developments on cluster ensemble.

The first step in cluster ensemble is to obtain the initial partitions. Many methods have been used to produce the base pool of the initial clustering results. In fact, they can be categorized into four subcategories, which are given below.
\begin{itemize}
\item Using different subsets or features of objects to produce different partitions \cite{strehl2002cluster}.
\item Projecting the data to subspaces, such as projecting to 1-dimension \cite{topchy2003combining}.
\item Changing the initialization or the number of clusters of a clustering algorithm \cite{christou2011coordination}.
\item Using different clustering algorithms to produce initial clustering results for combination \cite{dudoit2003bagging}.
\end{itemize}

The most challenging task in performing cluster ensemble is how to implement the second step. Generally speaking, it aims to find an appropriate consensus function which can improve the results of the individual clustering algorithm. Many works have been done to combine different partitions for obtaining consensus partition.

The relabeling and voting method \cite{dimitriadou2001voting} is a direct way to obtain consensus partition. Relabeling is to solve the problem of correspondence between two clusterings, then a voting mechanism is adopted to determine the final partition. In the process of relabeling, all the ensemble members must be relabeled based on a single reference partition which can be chosen from initial partitions or be generated by a single clustering algorithm. Unfortunately, using relabel to solve correspondence can make unsupervised cluster ensemble difficult.
Fischer and Buhmann \cite{fischer2003bagging} proposed a modified relabeling and voting method to combine the ensemble members. The correspondence problem between clusterings is solved through a maximum-likelihood problem employing the Hungarian algorithm. After relabeling, the cluster membership of each object can be assigned by a voting procedure. The simplicity and easy to implement are the advantages of these methods. However, the time complexity of the Hungarian algorithm is $O(k^3)$ which results in the high computational cost in combing different partitions, where $k$ is the number of clusters in consensus partition. Besides, in their works, the number of clusters in each clustering is assumed to be same and equals to the number of clusters in the final consensus partition which is a strong restriction to solve the problem of cluster ensemble.

Unlike the relabeling and voting methods, co-association methods need not consider the label correspondence between clusterings. It estimates the pairwise similarity among data points by counting the number of clusters shared by two objects in all partitions. The used co-association matrix is defined as
$$S_{ij}=S(x_i,x_j)=\frac{1}{M}\sum_{m=1}^M \delta(P^m(x_i),P^m(x_j)),$$
where
$$\delta (a,b) = \left\{ \begin{array}{l}
1{\kern 1pt} {\kern 1pt} ,{\kern 1pt} {\kern 1pt} {\kern 1pt} {\kern 1pt} {\kern 1pt} {\kern 1pt} {\kern 1pt} {\kern 1pt} {\kern 1pt} {\kern 1pt} if{\kern 1pt} {\kern 1pt} {\kern 1pt} {\kern 1pt} {\kern 1pt} {\kern 1pt} {\kern 1pt} {\kern 1pt} a = b\\
0{\kern 1pt} {\kern 1pt} ,{\kern 1pt} {\kern 1pt} {\kern 1pt} {\kern 1pt} {\kern 1pt} {\kern 1pt} {\kern 1pt} {\kern 1pt} {\kern 1pt} {\kern 1pt} otherwise
\end{array}, \right.$$
and $P^m(x_i)$ denotes the label assigned to $x_i$ in the $m$-th partition. Note that the values of the matrix $S$ ranges from 0 to 1, and the matrix $S$ can be regarded as a similarity matrix between a pair of objects in dataset. The co-association matrix can be used to recluster the data points by employing any similarity-based clustering algorithm \cite{fred2002data,ayad2003finding,topchy2004adaptive,fred2005combining}.
In general, "cutting" similarity matrix at a fixed threshold \cite{fred2001finding} or employing a single link algorithm and cutting the correspondence dendrogram at a certain similarity degree \cite{kuncheva2004using} is a common method to obtain a final partition. Considering that the traditional co-association matrix utilizes only cluster-data point relations, Iam-On et al. in their work proposed a link-based approach by adding the information among clusters in a refined co-association matrix \cite{iam2011link}. Although Iam-On's work have given a better performance than previous co-association matrix based methods, it only uses the similarity of clusters in one initial partition, and leads to a refined co-association matrix with much redundant information. To alleviate this problem, Hao et al. further proposed to refine the similarity matrix based on the similarity of clusters among all the initial partitions for reducing redundant information \cite{hao2015improved}.

Many algorithms about cluster ensemble make use of graph or hypergraph based representations to obtain the final consensus partition. The idea of these algorithms is to construct a graph or hypergraph based on ensemble members, and then employ graph partitioning algorithms to solve the cluster ensemble problem.
Strehl and Ghosh \cite{strehl2003cluster} proposed three graph based models to combine multiple partitions. The Cluster based Similarity Partitioning Algorithm (CSPA) uses the co-association matrix to generate a graph that the vertices are objects in a given dataset and the edge weights between vertices are the similarity degree of them. After that, it uses the graph partition package METIS to partition the graph. The second model, Hypergraph Partitioning Algorithm (HGPA), constructs a hypeygraph using the clusters of ensemble members and all hyperedges are considered to have the same weight. Then the hypergraph partitioning package is used to obtain a consensus partition. Furthermore, Meta Clustering Algorithm (MCLA) uses the hypergraph grouping and collapsing operations which can give a soft assignment for each object in the dataset. Fern and Brodley presented a bipartite graph partitioning algorithm \cite{fern2004solving} which solves the cluster ensemble problem by reducing it to a graph partitioning problem. This method uses both objects and clusters in the initial partitions to construct a bipartite graph and partitions the graph by a traditional graph partitioning technique.

Topchy et al. \cite{topchy2005clustering} defined category utility function to measure the similarity between two partitions. For two partitions $P^i=\{C_1^i,C_2^i,...,C_{k_i}^i\}$ and $P^j=\{C_1^j,C_2^j,...,C_{k_j}^j\}$, the category utility function is given below:
$$ U({P^i},{P^j}) = \sum\limits_{m = 1}^{{k_i}} {\rho (C_m^i)\sum\limits_{r = 1}^{{k_j}} {\rho (C_r^j|C_m^i) - \sum\limits_{r = 1}^{{k_j}} {\rho (C_r^j)} } }, $$
where $\rho (C_m^i) = \frac{{\left| {C_m^i} \right|}}{n}$, $\rho (C_r^j) = \frac{{\left| {C_r^j} \right|}}{n}$ and $\rho (C_r^j|C_m^i) = \frac{{\left| {C_r^j \cap C_m^i} \right|}}{{\left| {C_m^i} \right|}}$. Then the corresponding consensus function is obtained by solving the following optimization problem.
$${P^*} = \arg \mathop {\max }\limits_{P \in {\bf{P}}} \sum\limits_{i = 1}^M {U(P,{P^i})}. $$

Wu et al. \cite{wu2013theoretic} used a K-means-based Consensus Clustering (KCC) utility function to combine different clustering results. Analoui and Sadighian \cite{analoui2007solving} have presented a probabilistic model by a finite mixture of multinomial distributions in a space of clustering, where the consensus partition is obtained as the solution of a maximum likelihood estimation problem. Besides these, Alizadeh proposed a cluster ensemble selection mechanism by using a new cluster stability measure which was named Alizadeh-Parvin-Moshki-Minaei criterion (APMM) \cite{alizadeh2014cluster}. The framework of this work is to select as ensemble members a part of the initial partitions which satisfy APMM, and then find the final solution by using a co-association matrix based method.

Christou \cite{christou2011coordination} proposed the EXAct Method-based Cluster Ensemble (EXAMCE) algorithm based on the set partition problem which is direct to optimize the original objective function of cluster. It restricts the indictor matrix in the set partition problem consisting of the clusters in the base clustering results and converts it into a cluster ensemble problem.
EXAMCE can guarantee its optimal solution is at least as good as any clustering results in the base clusterings. However, it just modifies the indictor matrix and overlooks the structural information underlying in the initial partitions. Besides, it neglects the discriminative information lying in the raw data.

\section{Our proposed structure cluster ensemble}

In the following subsection, we first present the set partition problem. Then we formulate the cluster ensemble as a manifold regularized set-covering problem with a discriminative constraint, followed by our proposed SCE algorithm.

\subsection{Set partition problem}
As we all know, data clustering is to divide the data into $k$ clusters, which is equivalent to choose $k$ subsets from the data set such that any two subsets have no intersection and the union of all the subsets is the original data set.
Merle et al. \cite{du1999interior} looked on the MSSC problem as a set partition one and formulated it as
\begin{equation}
\begin{array}{l}
{\kern 1pt} \mathop {\min }\limits_x \sum\limits_{i = 1}^N {{c_i}{x_i}} \\
s.t.{\kern 1pt} {\kern 1pt} {\kern 1pt} {\kern 1pt} {\kern 1pt} {\kern 1pt} \left\{ \begin{array}{l}
{\kern 1pt} {\kern 1pt} {\kern 1pt} Ax = e,\\
{\kern 1pt} {\kern 1pt} \sum\limits_{i = 1}^N {{x_i} = k}, \\
{\kern 1pt} {x_i} \in \{ 0,1\} ,i = 1...N,
\end{array} \right.
\end{array}
\label{eq:1}
\end{equation}
where $A=[a_1,a_2,...,a_N] \in R^{n \times N}$ is an indicator matrix with value 0 or 1, the columns of matrix $A$ indicate each subset of the raw data and the value 1 means that the point belongs to this cluster, $N=2^n-1$ represents the number of subsets of data except empty set, $e$ is an $n$-dimension vector of ones. Note that the binary set $x$ is the solution vector, and the value of the $i$-th element in $x$ equals to 1 if the subset corresponding to the $i$-th column of A is chosen at the best clustering result and 0 otherwise. $c_i=c(a_i)$ is the cost of the $i$-th subset in matrix A and its definition is given below:
$${c_i} = \sum\limits_{x \in {C_i}} {\left\| {x - {m_i}} \right\|},$$
where
$${m_i} = \frac{{\sum\nolimits_{x \in {C_i}} x }}{{\left| {{C_i}} \right|}},$$
and $C_i$ represents the $i$-th subset.

From the formulation (\ref{eq:1}), we can see that the objective function is direct to minimize the total cost of chosen clusters. The first constraint is to ensure each point belongs to one and only one cluster, and the second constraint represents the number of clusters is $k$. It is easy to verify that there is no influence on the optimal solution by changing the equality constraints in the formulation (\ref{eq:1}) to inequalities:
\begin{equation}
Ax \geq e,
\label{eq:2}
\end{equation}
and
\begin{equation}
\sum_{i=1}^N x_i \leq k.
\label{eq:3}
\end{equation}
Next we give a simple explanation of the invariance property of the solution. For a solution of the set partition problem, if some rows are satisfied as $Ax>e$, it means that the data point corresponding to the row belongs to more than one cluster (denoted by $C_1$ and $C_2$ for clarity). Then we can obtain a solution with a more lower objective function value by replacing $C_1$ with a cluster containing all points of $C_1$ expect the point corresponding to the row. Note that the solution must exist since the indicator matrix $A$ contains total clusters.
If any solution has $m$ clusters and $m$ is lesser than $k$, we can choose $k-m$ points randomly from the data and let these points form $k-m$ new single point clusters, by which a better solution can be produced. So the optimum solution always satisfies $Ax=e$ and $\sum_{i=1}^N x_i = k$.
Note that the solutions are not always the same if the indicator matrix $A$ does not contain the total subsets of the data.

The formula of the set partition problem is a 0-1 integer programming which can be inverted into the general linear programming problem by relaxing the variable $x$. Then we can use many optimization algorithms for linear programming to solve it and obtain the relaxed solution which needs to be fixed to 0 or 1. However, when the number of data is
big, the size of the matrix $A$ increases exponentially. It is impossible to store the matrix $A$, not to mention obtaining the optimal solution. Based on the set partition problem, we present a set-covering problem and propose a manifold regularized set-covering method named Structure Cluster Ensemble (SCE) for clustering. Next we will give the formulation of the set-covering problem.

\subsection{Problem Formulation}
Given some initial clustering results, we aim to perform cluster ensemble to improve the performance of clustering. However, generally speaking, the given initial clusters only represent a part of subsets of the original data. So we cannot directly employ the optimization formulation of the above set partition problem to perform cluster ensemble.  Fortunately, the formulation can be modified to serve for cluster ensemble. In fact, we can perform cluster ensemble by solving a special set-covering problem and further performing the subsequent postprocessing. As we all know, given a set of elements (called the universe) and a set $S$ of $N$ sets whose union equals the universe, the general \textbf{set covering} problem is to identify the smallest subset of $S$ whose union equals the universe. If the set of all the initial clusters is looked on as the set $S$, and the $k$ final clusters as the smallest subset of $S$, then we can obtain a special set-covering problem with a constraint on the number of subsets of $S$.

As described in the previous subsection, the optimal solution to the set partition problem does not change if the first equality is relaxed to inequality. Furthermore, if the clusters $C_i$ and $C_j$ are similar, the corresponding values $x_i$ and $x_j$ in the solution $x$ should be the same. For the clusters $C_i$ and $C_j$, we use their cluster centers to represent them, and consequently employ heat kernel to calculate their similarity $A_{ij}$ defined by
$${A_{ij}} = {e^{ - \frac{{{{\left\| {{m_i} - {m_j}} \right\|}^2}}}{\sigma }}},$$
where $m_i$ and $m_j$ represent the cluster center of $C_i$ and $C_j$ respectively, and $\sigma$ is a constant. With the above defined similarity $A_{ij}$, we can deduce the following formula
$$\begin{array}{l}
\begin{split}
J &= \frac{1}{2}\sum\limits_{i,j = 1}^q {{{\left\| {{x_i} - {x_j}} \right\|}^2}{A_{ij}}} \\
   &= \sum\limits_{i = 1}^q {x_i^2{D_{ii}} - \sum\limits_{i,j = 1}^q {{x_i}{x_j}{A_{ij}}} } \\
   &= {x^T}Dx - {x^T}Ax\\
   &= {x^T}Lx,
\end{split}
\end{array}$$
where $A = {\left[ {{A_{ij}}} \right]_{i,j = 1,..,q}}$ is the adjacent matrix, $q$ is the number of initial clusters, $D$ is a diagonal matrix whose each entry is row (or column, because $A$ is symmetric matrix) sum of $A$, denoted as ${D_{ii}} = \sum\nolimits_j {{A_{ij}}} $, and $L = D - A$, which is called graph Laplacian \cite{qi2015successive,gong2015deformed}. It has been well studied in manifold learning that the graph Laplacian can effectively capture the manifold structure lying in the data \cite{cai2011graph,guan2011manifold}. Therefore, the similar clusters may have the same values in the solution $x$ by minimizing $J$. Incorporating this manifold regularization into the set partition problem leads to our structure cluster ensemble method.
Additionally, we take the discriminative information among the raw data into account with a limitation on the inter-class distance which follows the cluster criterion of keeping high dispersion degree
between clusters. So we can obtain a manifold regularized set-covering problem with a discriminative constraint. Our objective function is formularized as
\begin{equation}
\begin{array}{l}
(SCE)\mathop {\min }\limits_x \sum\limits_{i = 1}^q {c({{[{A_B}]}_i}){x_i} + \beta({x^T}Lx}) \\
s.t.{\kern 1pt} {\kern 1pt} {\kern 1pt} {\kern 1pt} {\kern 1pt} {\kern 1pt} {\kern 1pt} {\kern 1pt} \left\{ \begin{array}{l}
{\kern 1pt} {A_B}x \ge e,\\
\sum\limits_{i = 1}^q {{x_i} = k}, \\
\sum\limits_{i = 1}^q {{d_i}{x_i} \ge \eta }, \\
{x_i} \in \{ 0,1\} ,i = 1...q,
\end{array} \right.
\end{array}
\label{eq:4}
\end{equation}
where $A_B$ contains only the clusters that have been obtained by the initial clustering algorithms, $c({{[{A_B}]}_i})$ is the cost of the $i$-th column in the matrix $A_B$, $\beta$ is a weight factor to balance the effects of two terms, ${d_i}$ represents the minimum
distance of the $i$-th cluster in matrix $A_B$ with other clusters in the initial clustering results, and $\eta$ is the cut-off factor to limit the sum of the minimum distance of clusters in the solution.

Note that $x\in {R^q}$ is the solution vector including binary values 0 and 1. The value of the $i$-th element in $x$ equals to 1 if the corresponding cluster is chosen and 0 otherwise. In order to keep the similar clusters with the same value in the solution $x$, we construct a Laplacian matrix $L$ measured by using the similarity between the initial cluster centers. Furthermore, the Laplacian regularization is added to the objective function for maintaining the manifold structure of the raw data. In other words, the second term in equation (4) is a Laplacian regularization term which is designed for capturing the structure information.
Besides, the first constraint may result in overlapping clusters in the final solution. To deal with this situation, we can use any overlaps removal techniques to ensure each point belongs to only one cluster. In this paper, we assign the point appearing in more than one cluster to their nearest cluster center. The second constraint means that the number of final clusters is $k$.  In the third constraint, we aim to make sure the dispersion degree of final clusters is not too small. It can be considered as a discriminative constraint. For simplicity, we adopt the summation of minimum distances between the initial clusters instead of final clusters to represent the dispersion degree of the final clusters. In this way, this discriminative constraint can be simplified as a linear one, which will reduce the computational complexity of solving the set-covering problem. The following part gives the framework of our SCE algorithm.

\subsection{SCE algorithm}
In this subsection, we will present our proposed Structure Cluster Ensemble (SCE) algorithm to perform data partitioning. Particularly, we first use the clustering algorithm MMDA to obtain the initial clusters, and then solve our proposed set-covering problem (\ref{eq:4}). Finally, we perform the postprocessing to adjust the solution for satisfying the demand of data partitioning. These steps will be described in more detail below.
\subsubsection{Initial Clustering}
\begin{figure}[!t]
\begin{center}
\includegraphics[width=3.6in]{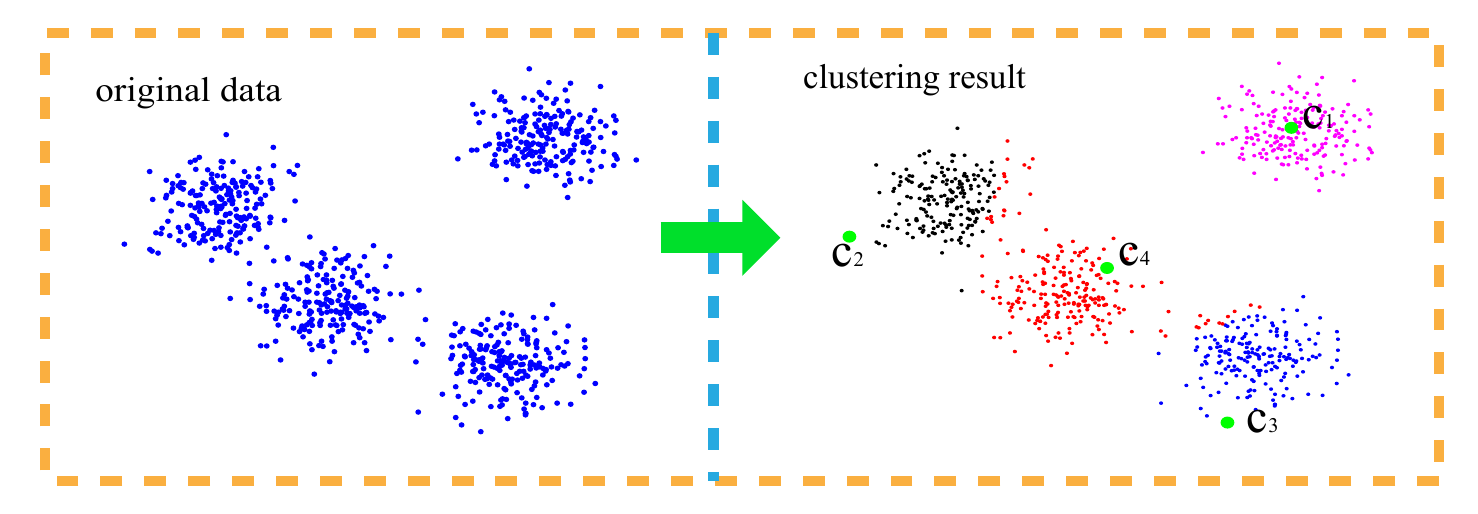}
\end{center}
\caption{An illustration of using MMDA  to cluster the four-gauss data. $c_1 \sim c_4$ represent different cluster centers. Different colors mean different clusters.}
\label{li2}
\end{figure}
At the first step of our SCE algorithm, we employ MMDA, a common method to generate the initial cluster centers for improving the performance of the traditional K-means algorithm, to produce the individual partitions. MMDA can be regarded as a week clustering algorithm and has low time complexity which can save much time during generating the initial clustering results. In order to increase the diversity of the initial partitions, we set different number of clusters that are close to the final number of clusters for different clustering results. The main steps of MMDA are given below:
\begin{enumerate}
  \item Set the number of clusters $k$ and choose one point from dataset $X$ randomly as the first center $c_1$.
  \item For each point, we find its nearest neighbor from the centers we have already chosen and record the distance between them denoted by $D(x)$.
  \item Choose the center $c_i$, selecting ${c_i} = \mathop {\arg \max }\limits_{x \in X} D(x)$.
  \item Repeat Step 2 and 3 until the number of centers is $k$.
  \item Find the nearest neighbor from the set $\{c_1,c_2,...,c_{k}\}$ for each $x\in X$ and update cluster centers. A clustering result $P=\{C_1,C_2,...,C_k\}$ is obtained.
\end{enumerate}
Then we can obtain different clustering results by changing the value of $k$.

From the above steps, we can see that the MMDA algorithm aims to keep cluster centers with large distance which usually leads to the cluster centers locating in the edge of the data. Fig. \ref{li2} gives a demonstration of MMDA dealing with four-gauss data set. Although the MMDA algorithm can not give a good performance on a single clustering result, it has the low time complexity $O(nk)$ and we can use our cluster ensemble method to produce a final clustering result with high quality.

\subsubsection{Ensemble Method}
\begin{figure*}[htpb]
\begin{center}
\includegraphics[width=5.9in]{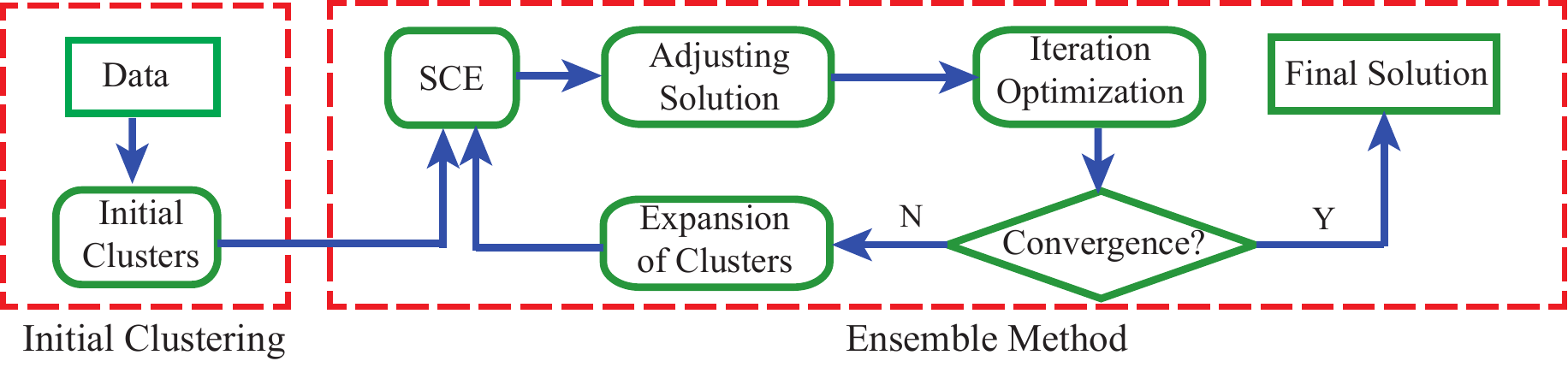}
\end{center}
\caption{The flow diagram of our SCE algorithm. The left part is to obtain initial clustering results. And the right part combines these initial clustering results and finds a better final result.}
\label{li3}
\end{figure*}

After obtaining the initial clustering results, we formulate the set-covering problem. During solving the set-covering problem, a 0-1 integer quadratic programming problem, we relax the integer constraint of the solution vector with ${x_i} \in [0,1]$ and use the interior point convex algorithm to solve this problem.

Due to that the solution of the formulation (\ref{eq:4}) may contain overlapping clusters, we execute the step of \emph{Adjusting solution} which is given below to obtain a feasible clustering result $P$. Considering that the result $P$ is often not a local minimum solution, iteration optimization method is employed to optimize the clustering result $P$ and obtain a better performance ${P^{'}}$.
Then we expand the clusters of $P$ and ${P^{'}}$ into a clusters set ${P^{''}}$ and add the clusters of $P$, ${P^{'}}$ and ${P^{''}}$ to the indicator matrix $A_B$.
Afterwards, we can update the set-covering problem and obtain a more efficient clustering result. Repeat the above steps until the relative error is less than the threshold value. The whole framework of our SCE algorithm is presented in Fig. \ref{li3} which shows the main procedure of \emph{Ensemble Method}. The following gives the detailed description of \emph{Adjusting solution, Iteration optimization and Expansion of the clusters}.
\begin{algorithm}[!t]
\caption{Structural cluster ensemble}
\label{alg:1}
\textbf{Input:}

\ $X$, the data set

\ $k$, the number of clusters

\textbf{Output:}

\ $P^*$, the final solution

\textbf{Processing:}

\begin{enumerate}
\item Use the base clustering algorithm MMDA to produce the initial clustering results and obtain the indicator matrix $A_B$
\item Construct and solve the objective function with a side constraint:
\begin{equation}
\begin{array}{l}
(SCE)\mathop {\min }\limits_x \sum\limits_{i = 1}^q {c({{[{A_B}]}_i}){x_i} + \beta({x^T}Lx}) \\
s.t.{\kern 1pt} {\kern 1pt} {\kern 1pt} {\kern 1pt} {\kern 1pt} {\kern 1pt} {\kern 1pt} {\kern 1pt} \left\{ \begin{array}{l}
{\kern 1pt} {A_B}x \ge e\\
\sum\limits_{i = 1}^q {{x_i} = k} \\
\sum\limits_{i = 1}^q {{d_i}{x_i} \ge \eta } \\
{x_i} \in [0,1] ,i = 1...q
\end{array} \right.
\end{array}
\label{eq:5}
\end{equation}
\item For the solution $x$ of SCE, we use the step of \emph{Adjust solution} to obtain a feasible  solution $P$
\item Use \emph{iteration optimization} to produce a new local minimum solution $P{'}$
\item Call \emph{expansion of the clusters} to expand the clusters of $P$ and $P{'}$ for obtaining a new clusters set $P{''}$
\item Add all clusters of the union $P \cup {P^{'}} \cup {P^{''}}$ to the indicator matrix $A_B$
\item Repeat Steps 2-6 until the solution satisfies the convergence criteria
\end{enumerate}
\end{algorithm}

\emph{Adjusting solution:} For the solution $\tilde x$, we rank the elements of the vector $\tilde x$ in descending order and set the first $k$ values to one, others to zero. After that, we can obtain $k$ clusters which may have overlaps. To solve this problem, we assign the point appearing in more than one cluster to their nearest cluster centers and make the unlabel points belong to the clusters with nearest distance. And then we can obtain a feasible solution $P$.

\emph{Iteration optimization:} Firstly, we regard the solution $P$ as the initial cluster centers and use the traditional K-means algorithm to cluster the original data. Then iteration optimization method, a single-move steepest descent algorithm \cite{hansen2001j}, is applied for the solution produced by K-means and obtain a local minimum solution ${P^{'}}$.

\emph{Expansion of the clusters:} For each cluster $C_i$ in the solution $P$ and $P^{'}$, we make a small disturbance for $C_i$ by adding a point that is not in cluster and has the smallest distance from the cluster center and removing a point that is the member of cluster and has the largest distance from the cluster center. We then create a new set ${P^{''}}$ containing partial subsets of the data: $ S_1^ +  \subset S_2^ +  \subset ... \subset S_\omega ^ + $, $S_1^ -  \supset S_2^ -  \supset ... \supset S_\omega ^ - $, where $ S_i^ + $ contains all points in the cluster $C_i$ plus up to the $i$-th nonmember nearest neighbor from the cluster center and $ S_i^ - $ contains all members in the cluster $C_i$ except up to the $i$-th farthest point from the cluster center. Suggested by Christou's work \cite{christou2011coordination}, we set the value of $\omega$ to 10.

Algorithm \ref{alg:1} summarizes the whole steps of the SCE algorithm. For the raw data $X$, we first utilize MMDA to produce initial clustering results with different number of clusters. Then we solve the set-covering problem and use a series of optimizing procedures to obtain a local minimum solution. Finally, we expand the indicator matrix and acquire the final solution in an iterative approach. Note that our proposed method converges quickly, and in many cases, it can reach the local optimization solution after iterations.

\section{Experimental results}

In this section, we will apply our proposed SCE algorithm on ten datasets consisting of five low-dimensional datasets and five high-dimensional image datasets to demonstrate its effectiveness. We first describe the experimental settings in the following subsection.

\subsection{Experimental Settings}
\subsubsection{Data Set}

To effectively verify our proposed method, we  in our experiments use ten datasets presented in Table I.
\begin{table}[t]
\renewcommand{\arraystretch}{1}
\caption{Data sets used in experiments}
\centering
\begin{tabular}{c|c|c|c}
\hline
Dataset & Clusters & Number & Dimension\\
\hline
\hline
Gauss60 & 60 & 1200 & 2\\
u1060 & N/A & 1060 & 2\\
pcb3038 & N/A & 3038 & 2\\
Wine & 3 & 178 & 13\\
SIFT & N/A & 400 & 128\\
USPS & 10 & 9298 & 256\\
FEI & 50 & 700 & 768\\
Yale & 15 & 165 & 1024\\
PIE & 68 & 11554 & 1024\\
Olivettifaces & 40 & 400 & 4096\\
\hline
\end{tabular}
\label{table:1}
\end{table}

\begin{figure}[t]
\begin{center}
\includegraphics[width=3in]{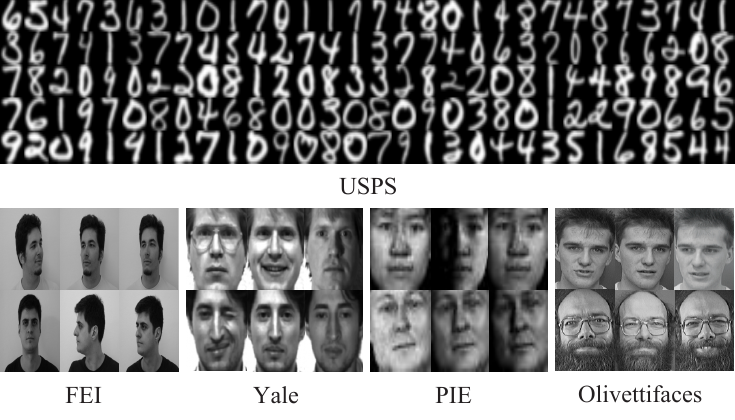}
\end{center}
\caption{Some examples of handwritten digit images and  face databases.}
\label{li4}
\end{figure}

Gauss60 is a synthetic data set with 60 clusters. u1060 and pcb3038 datasets \cite{laszlo2006genetic,pacheco2005scatter} were used in the experiments of Christou's work \cite{christou2011coordination} comparing with other clustering algorithms. We adopt these two datasets for comparing with the EXAMCE algorithm and any other popular clustering algorithms using their available datasets. Wine is a popular data set for cluster analysis with three clusters. SIFT is part of sift descriptors extracted from Oxford building used in image retrieval \cite{philbin2007object,jegou2014triangulation,philbin2008lost}.

USPS is  a handwritten digit database. FEI, Yale, PIE and Olivettifaces are face databases which are widespread benchmarks for algorithms in machine learning. We use these five high dimensional image datasets to verify the effectiveness of our SCE algorithm in capturing the manifold structure of dataset. USPS contains 9298 16$\times$16 handwritten digit images in total. FEI data set has 700 pictures of 50 individuals, and each person has 14 images. Yale face \cite{xu2015topology} database contains 165 gray images of 15 different human faces with different pose, angle and illumination. PIE database \cite{7072521} has faces of 68 people with more than ten thousand images which is a large dataset for clustering analysis. Olivettifaces \cite{rodriguez2014clustering} has total 400 grayscale faces images with 10 individuals.
Fig. \ref{li4} is partial images of these five datasets.\
\subsubsection{Evaluation Criterion}
In our experiments, we use the Minimum Sum-of-Squares Clustering (MSSC) criterion to evaluate our method. So the cost function $c_i$ in formulation (\ref{eq:4}) is the squares error of the $i$-th cluster in matrix $A_B$.
In fact, the MSSC value is the sum-of-squared distances from each entity to the centroid of the final cluster to which it belongs. The formulation of the MSSC criterion is given below.
$$\begin{array}{l}
\min \mathop \sum\limits_{i = 1}^k {\sum\limits_{x \in {C_i}} {{{\left\| {x - {m_i}} \right\|}_2^2}} } \\
\end{array}$$
where
$$\begin{array}{l}
{m_i} = \frac{{\sum\nolimits_{x \in {C_i}} x }}{{\left| {{C_i}} \right|}},{\kern 1pt} {\kern 1pt} {\kern 1pt} {\kern 1pt} i = 1,...,k,{\kern 1pt} \\
\mathop  \cup \limits_{i = 1}^k {C_i} = X,\\
{C_i} \cap {C_j} = 0,{\kern 1pt} {\kern 1pt} {\kern 1pt} \forall i \ne j.
\end{array}$$
and $C=\{C_1,C_2,...,C_k\}$ is the final clusters. Note that, the smaller the MSSC value is, the better performance the ensemble  cluster method achieves.
\begin{figure}[!t]
\centering
\subfigure[Wine]{\includegraphics[width=4.2cm]{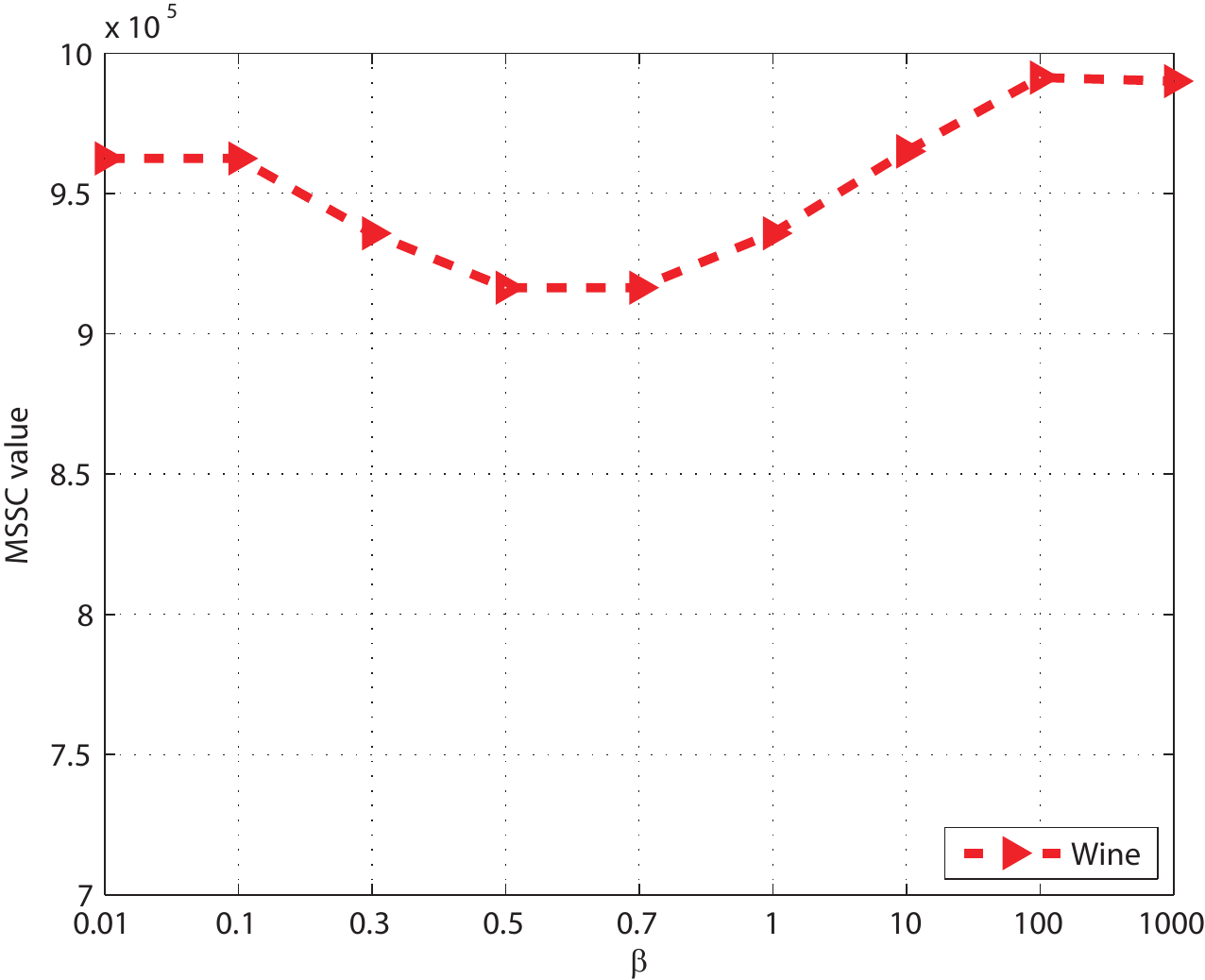}%
\label{li5a}}
\hfil
\subfigure[Yale]{\includegraphics[width=4.2cm]{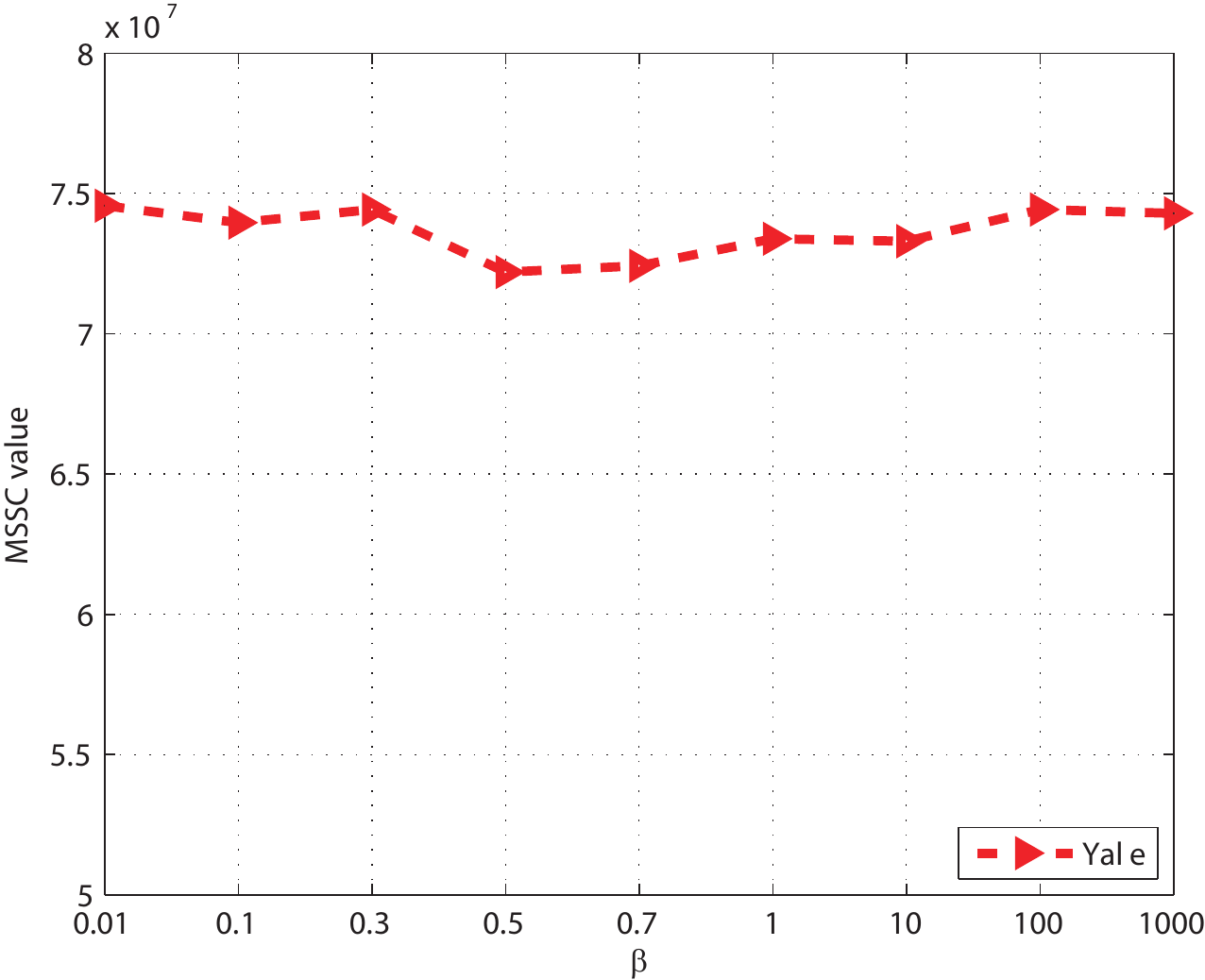}%
\label{li5b}}
\caption{The performance of our method versus parameter $\beta$.}
\label{li5}
\end{figure}

\subsubsection{Parameter Setting}
\begin{table*}[!t]
\begin{center}
\renewcommand{\arraystretch}{1.1}
\caption{Evaluation of the parameter $M$ on Gauss60 and Olivettifaces dataset\label{table:2}}
\centering
\begin{tabular}{c|c|c|c|c|c|c|c}
\hline
Dataset & \backslashbox {k}{M} & 5 & 10 & 20 & 30 & 40 & 50\\
\hline\hline
\multirow{5}{*}{Guass60}
& 5  & \textbf{1.3929E+05} &\textbf{ 1.3929E+05} & \textbf{1.3929E+05} & \textbf{1.3929E+05} & \textbf{1.3929E+05} & \textbf{1.3929E+05} \\
\cline{2-8}
& 10 & 5.1414E+04 & 5.1410E+04 & \textbf{5.1409E+04} & \textbf{5.1409E+04} & \textbf{5.1409E+04} & \textbf{5.1409E+04} \\
\cline{2-8}
& 15 & 2.7946E+04 & 2.7948E+04 & \textbf{2.7944E+04} & \textbf{2.7944E+04} & 2.7948E+04 & 2.7947E+04 \\
\cline{2-8}
& 20 & 1.8642E+04 & 1.8541E+04 & 1.9464E+04 & \textbf{1.8540E+04} & 1.8541E+04 & 1.8646E+04 \\
\cline{2-8}
& 30 & 1.0233E+04 & 1.0199E+04 & 1.0479E+04 & \textbf{1.0196E+04} & 1.0585E+04 & \textbf{1.0196E+04} \\
\cline{2-8}
& 40 & 6.4640E+03 & \textbf{6.3864E+03} & 6.9064E+03 & 6.3866E+03 & 7.2306E+03 & 7.4953E+03 \\
\hline
\multirow{5}{*}{Olivettifaces}
& 5  & 1.0444E+08 & \textbf{1.0325E+08} & \textbf{1.0325E+08} & \textbf{1.0325E+08} & \textbf{1.0325E+08} & \textbf{1.0325E+08} \\
\cline{2-8}
& 10 & \textbf{2.0704E+08} & \textbf{2.0704E+08} & 2.0725E+08 & \textbf{2.0704E+08} & \textbf{2.0704E+08} & \textbf{2.0704E+08} \\
\cline{2-8}
& 15 & 2.8421E+08 & 2.8051E+08 & 2.8057E+08 & \textbf{2.8030E+08} & \textbf{2.8030E+08} & \textbf{2.8030E+08} \\
\cline{2-8}
& 20 & 3.8099E+08 & 3.7498E+08 & 3.7452E+08 & 3.7318E+08 & \textbf{3.7301E+08} & 3.7550E+08 \\
\cline{2-8}
& 30 & 5.2339E+08 & 5.1568E+08 & 5.2197E+08 & \textbf{5.1464E+08} & 5.1760E+08 & 5.2375E+08 \\
\cline{2-8}
& 40 & 7.0093E+08 & 6.8915E+08 & 6.8963E+08 & \textbf{6.8572E+08} & 6.9129E+08 & 7.0088E+08 \\
\hline
\end{tabular}
\end{center}
\end{table*}

To increase the diversity in initial clustering results, we run the MMDA algorithm with different number of clusters $K$, where $K=k-\tau,...,k-1,k,k+1,...,k+\tau$, and $\tau$ is set as $\left\lfloor {\frac{k}{{10}}} \right\rfloor $ suggested by Christou's work \cite{christou2011coordination}.

In order to find a suitable $M$ (the number of initial clustering results), we conduct a parameter experiment on Gauss60 and Olivettifaces datasets. Gauss60 is a synthetic dataset which is common used in cluster analysis. Olivettifaces is a real one with the highest dimension of our datasets. We use these two representative datasets to find a reasonable $M$. Table \ref{table:2} shows the corresponding experiment results that the first row represents the different value of $M$. As we can see from the table, when the value of $k$ is 5 or 10, different $M$ have almost the same performance. But along with the increasing of $k$, setting $M=30$ can usually give a better result and have a stable performance. Considering the time complexity of solving SCE problem and the performance of the final clustering result, we obtain 30 individual clustering results in all our experiments.

Our objective function has two essential parameters: $\beta$ and $\eta$. The parameter $\beta$ is a weight factor to balance the effects of two terms. In order to choose an appropriate value, we test the parameter $\beta$ on Wine and Yale data sets. The range of $\beta$ is chosen from $10^{-2}$ to $10^3$. The corresponding number of clusters is selected as 5. Fig. \ref{li5} gives the experimental results. Note that our method can obtain a satisfactory performance when $\beta$ belongs to $[0.5,0.7]$. For simplicity, we choose the average $\beta=0.6$ in all the experiments.

The parameter $\eta$ is a cut-off factor to keep high dispersion degree between clusters. In fact, $d_i$ in equation (\ref{eq:4}) represents the minimum distance of the $i$-th cluster in matrix $A_B$ with other clusters in the initial clustering results. We construct the third constraint in order to make the summation of their $d_i$ be larger than $\eta$. To estimate the value of $\eta$, we need to estimate the minimum distance of a random cluster with others. For solving this problem, we calculate the minimum value of inter-class distances in a single clustering result, denoted by $\theta$, and regard $\theta$ as the estimated value. Considering that the number of clusters is $k$, we can set $\eta$ having the same order of magnitudes with $k\theta$.
\begin{figure}[!t]
\centering
\subfigure[Wine]{\includegraphics[width=4.2cm]{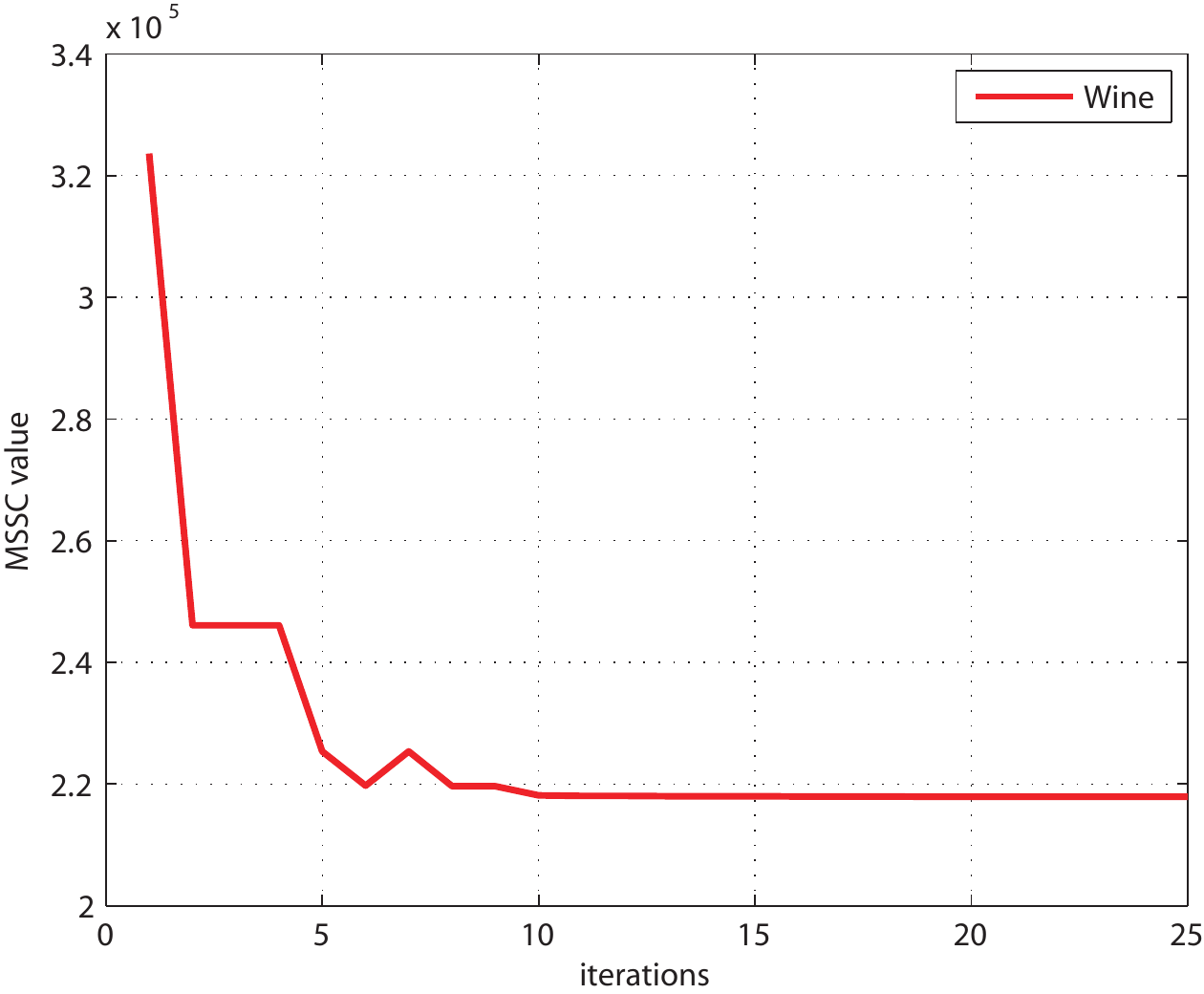}%
\label{li6a}}
\hfil
\subfigure[Yale]{\includegraphics[width=4.25cm]{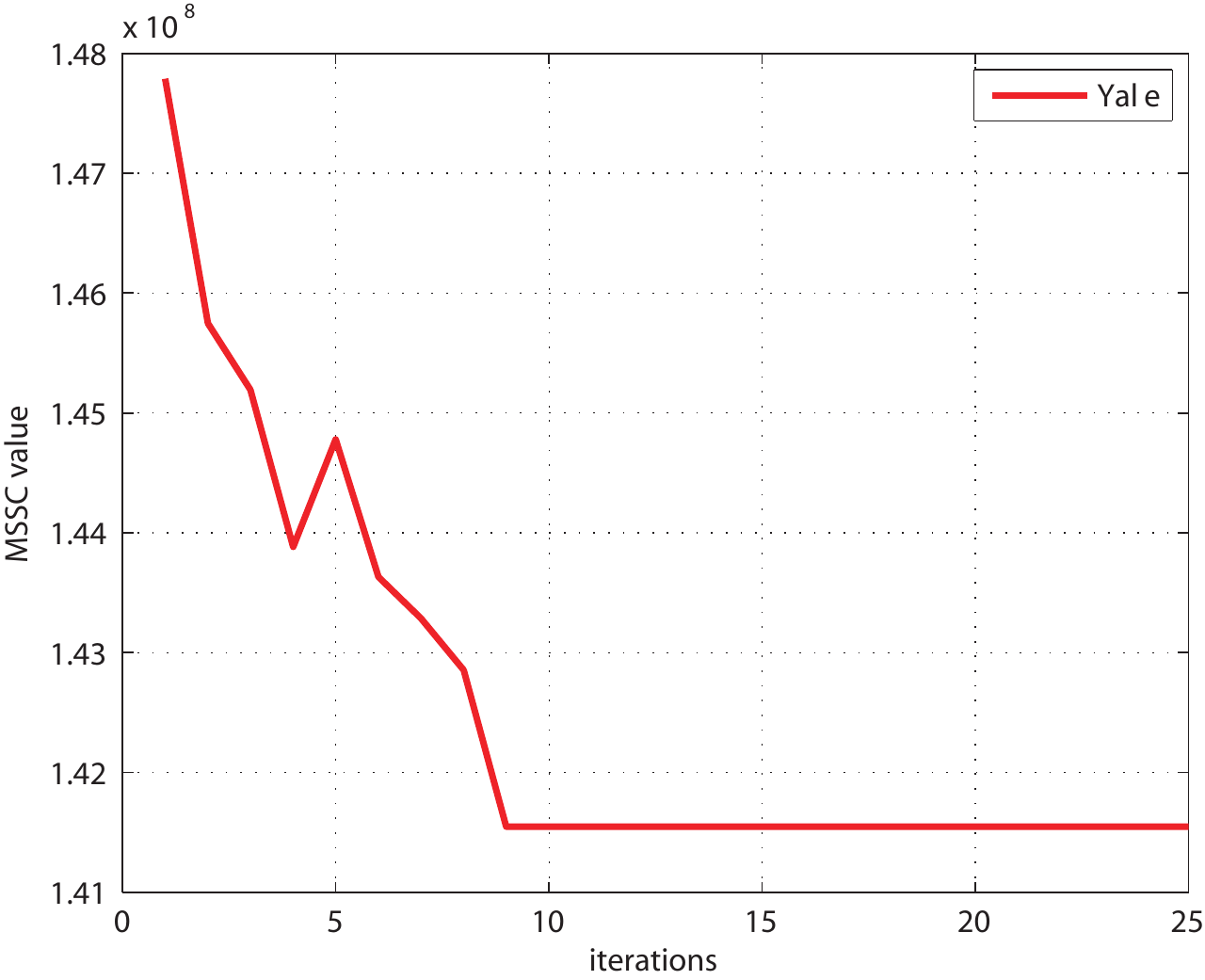}%
\label{li6b}}
\caption{The convergence experiment of our SCE method on Wine and Yale datasets.}
\label{li6}
\end{figure}

\begin{table*}[!t]
\begin{center}
\renewcommand{\arraystretch}{1}
\caption{The MSSC values of K-means and SCE on Gauss60 data set\label{table:7}}
\centering
\begin{tabular}{*{8}{c|} c}
\hline
k & 5 & 10 & 15 & 20 & 30 & 40 & 50 & 60\\
\hline\hline
K-means & \textbf{1.3929E+05} & 5.1419E+04 & 2.7951E+04 & 1.9259E+04 & 1.1085E+04 & 6.8560E+03 & 5.4918E+03 & 4.1259E+03\\
SCE & \textbf{1.3929E+05} & \textbf{5.1409E+04} & \textbf{2.7944E+04} & \textbf{1.8540E+04} & \textbf{1.0196E+04} & \textbf{6.3866E+03} & \textbf{4.5581E+03} & \textbf{3.8686E+03}\\
\hline
\end{tabular}
\end{center}
\end{table*}

\subsubsection{The convergence analysis of SCE}

We begin by stating why the solution by Algorithm 1 is a feasible solution to optimization problem in equation (\ref{eq:4}). From the optimization problem defined in equation (\ref{eq:4}), we can see it is an integer non-linear programming problem which is NP-hard and difficult to find the solution. In order to solve the equation (\ref{eq:4}) efficiently, we relax the integer constraint of the solution vector with $x_i\in[0,1]$. Although the solution of relaxation version is not a feasible solution to formula (\ref{eq:4}), the step of \emph{Adjusting solution} in Algorithm 1 can be used to obtain a feasible solution. For the solution of relaxation version, we first rank the elements of the solution in descending order and set the first $k$ values in which the corresponding $d_i$ are bigger than $\eta/k$ to one, others to zero. After that, a binary solution is obtained and the number of clusters is equal to $k$. Moreover, the summation of $d_i$ that we chose is no less than $\eta$. So the second, third and fourth constraints of the formula (\ref{eq:4}) are satisfied. For now, the clusters may have overlaps and not contain all the points, which mean some points belong to more than one cluster and some points have no label. In order to solve this problem, we assign the points appearing in more than one cluster to their nearest cluster and make the unassigned points belong to the cluster with nearest distance. Finally, the first constraint is also satisfied and a feasible solution of the formula (\ref{eq:4}) is obtained. So the solution produced by Algorithm 1 is the solution to optimization problem in equation (\ref{eq:4}).

\begin{table*}[t]
\begin{center}
\renewcommand{\arraystretch}{1}
\caption{The MSSC values of different algorithms on the data set u1060 and pcb3038\label{table:3}}
\centering
\begin{tabular}{*{8}{c|} c}
\hline
Dataset & k & J-Means & VNS+ & GA-QdTree & SS & DA & EXAMCE & SCE\\
\hline\hline
   \multirow{6}{*}{u1060}
   & 10 & 1.7564E+09 & \textbf{1.7548E+09} & 1.7559E+09 & \textbf{1.7548E+09} & 1.7565E+09 & \textbf{1.7548E+09} & \textbf{1.7548E+09}\\
   & 20 & 8.1895E+08 & \textbf{7.9179E+08} & \textbf{7.9179E+08} & \textbf{7.9179E+08} & 8.2483E+08 & \textbf{7.9179E+08} & \textbf{7.9179E+08}\\
   & 30 & 5.0141E+08 & \textbf{4.8125E+08} & 4.8155E+08 & \textbf{4.8125E+08} & 5.2961E+08 & 4.8137E+08 & 4.8131E+08\\
   & 50 & 2.6915E+08 & \textbf{2.5551E+08} & 2.5689E+08 & 2.5643E+08 & 3.1243E+08 & \textbf{2.5551E+08} & \textbf{2.5551E+08}\\
   & 60 & 2.0546E+08 & \textbf{1.9727E+08} & 1.9874E+08 & 1.9738E+08 & 2.4156E+08 & \textbf{1.9727E+08} & \textbf{1.9727E+08}\\
   & 70 & 1.6434E+08 & \textbf{1.5845E+08} & 1.5991E+08 & \textbf{1.5845E+08} & 2.0269E+08 & \textbf{1.5845E+08} & \textbf{1.5845E+08}\\
\hline\hline
   \multirow{5}{*}{pcb3038}
   & 10 & 5.6350E+08 & \textbf{5.6025E+08} & \textbf{5.6025E+08} & \textbf{5.6025E+08} & 5.7548E+08 & \textbf{5.6025E+08} & \textbf{5.6025E+08}\\
   & 20 & 2.6695E+08 & \textbf{2.6681E+08} & 2.6684E+08 & \textbf{2.6681E+08} & 2.6734E+08 & 2.6686E+08 & \textbf{2.6681E+08}\\
   & 30 & 1.7649E+08 & 1.7560E+08 & \textbf{1.7557E+08} & 1.7560E+08 & 1.8078E+08 & \textbf{1.7557E+08} & \textbf{1.7557E+08}\\
   & 40 & 1.2826E+08 & 1.2607E+08 & 1.2533E+08 & \textbf{1.2496E+08} & 1.3578E+08 & \textbf{1.2496E+08} & \textbf{1.2496E+08}\\
   & 50 & 1.0061E+08 & 9.8944E+07 & 9.8641E+07 & 9.8340E+07 & 1.0785E+08 & \textbf{9.8275E+07} & 9.9049E+07\\
\hline
\end{tabular}
\end{center}
\end{table*}

Next we will give the detailed convergence analysis. In Algorithm \ref{alg:1}, Step 2 uses the interior point convex algorithm to solve the equation (\ref{eq:5}) and a series of solutions converging to the global optimal solution can be obtained. Considering the resulting solution is not a feasible solution to problem (\ref{eq:4}), we use the \emph{Adjust solution} method in Step 3 to convert the elements in solution to integer (0 or 1) and obtains a feasible solution. Furthermore, Step 4 uses \emph{Iteration optimization} to optimize the feasible solution and consequently obtains a local minimum solution. Note that Step 2 and Step 4 can give a convergent solution. Though the solution of every iteration of Algorithm \ref{alg:1} is not monotone decreasing, a stable solution can be always obtained after iterations in our experiments. Fig. \ref{li6} gives the experiment results of convergence analysis.
The experiment is conducted on Wine and Yale data sets. The corresponding number of clusters is selected as 10. The horizontal ordinate represents the number of iteration and the vertical coordinate is the MSSC value. From Fig. \ref{li6}, we can see that the MSSC value is stable when the number of iteration is bigger than 10 on both of the data sets. In our experiments, Algorithm 1 will be stopped when the relative error of the MSSC values of successive two iterations is less than 0.5\% or the number of iteration is bigger than a given integer (50 is enough in our experiments).

\subsection{Performance and Comparisons}

\begin{figure}[t]
\begin{center}
\includegraphics[width=2.0in]{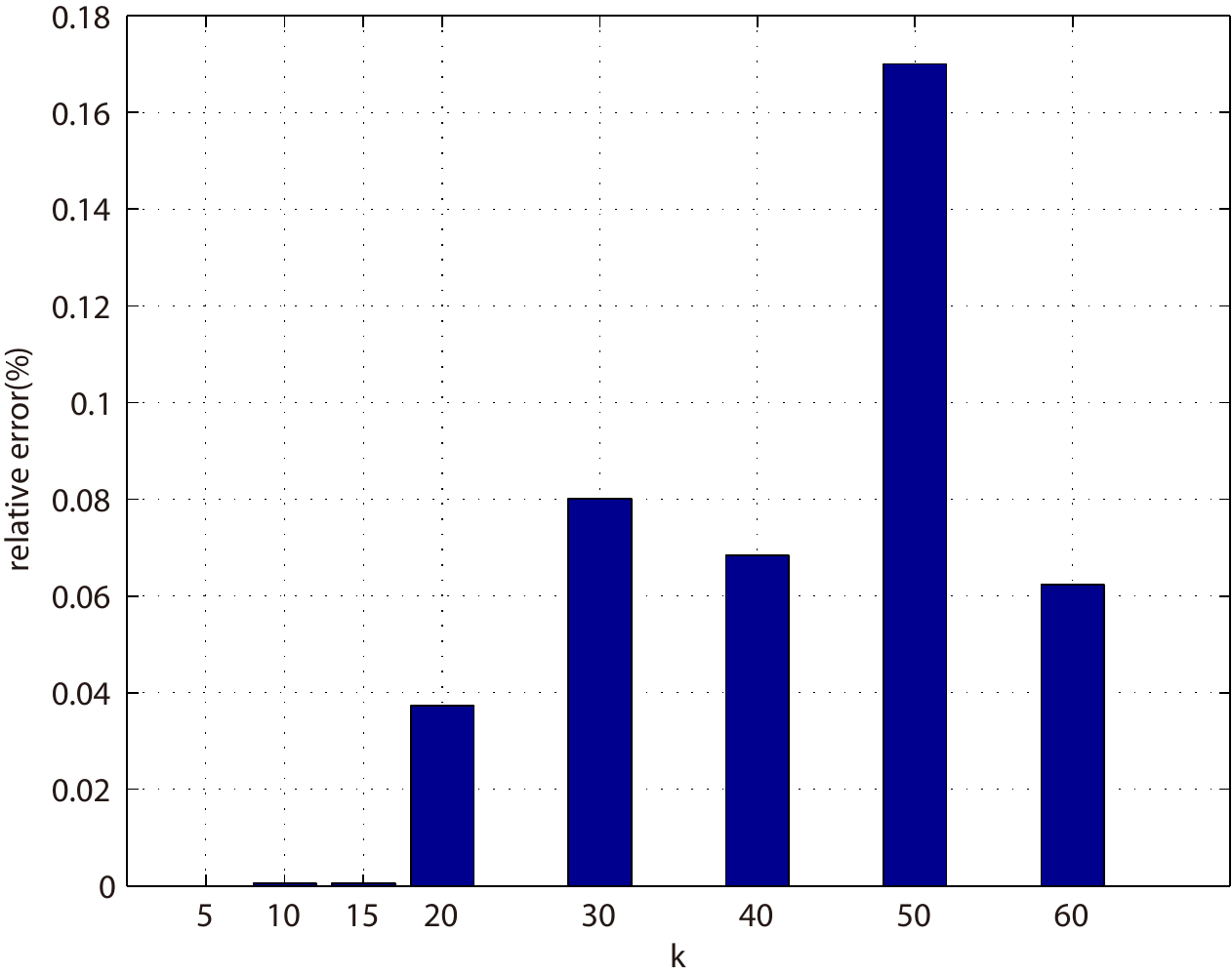}
\end{center}
\caption{The relative error between K-means and our method on Gauss60 data set.}
\label{li7}
\end{figure}

In this subsection, we first conduct a preliminary experiment on Gauss60 dataset for demonstrating the performance of our proposed method. Considering the traditional K-means method is a popular clustering algorithm in many clustering analysis task and is suitable for clustering Gaussian data, we also give the clustering results of K-means on Guass60. Table \ref{table:7} gives the corresponding MSSC values, and it shows that our method obtains the best clustering result for all the cases. Fig. \ref{li7} demonstrates the relative error of K-means with our method. The positive values in Fig. \ref{li7} represent the improvement of our method compared to K-means. From Fig. \ref{li7}, it can be obviously seen that our method achieves better performance than K-means, especially when $k$ belongs to [20,60].

To further demonstrate the effectiveness of our proposed SCE algorithm, we first compare it on the u1060 and pcb3038 datasets with six representative clustering algorithms, including J-Means, Variable Neighborhood Searching (VNS+), Genetic Algorithm Using Hyper-Quadtrees (GAUHQ), Scatter Search (SS), Deterministic Annealing (DA) and EXAMCE. Table \ref{table:3} shows their clustering results. J-Means and VNS+ were proposed by Hansen's work. The J-Means algorithm \cite{hansen2001j} is a local search heuristic for MSSC that adds a new cluster center with a non-centroid point in the data and deletes a cluster center closed to this point for increasing the probability of improving the performance of clustering result. Then K-Means, H-Means, or some other heuristic methods can be used to optimize the solution.
VNS \cite{mladenovic1997variable} is a simple and effective metaheuristic algorithm for combinatorial optimization following a local search routine. It always finds the neighborhood of the current solution and jumps from the current result to a new solution if and only if an improvement was made. The VNS+ algorithm replaces J-Means at the local search step with the J-Means+ heuristic. The GAUHQ algorithm \cite{laszlo2006genetic} uses the genetic algorithm to evolve centers in the K-means algorithm and represents the set of centers with a hyper-quadtree constructed on the data. It can find good initial centers and has ability to obtain the global optimum for data sets. SS \cite{pacheco2005scatter} can be regarded as a combinatorial method that combines subsets in the reference set and obtain new solutions. SS utilizes procedures based on different methods, such as Local Search, Tabu Search, GRASP or Path Relinking.
For small numbers of clusters, it always obtain a good performance. The DA \cite{rose1998deterministic} algorithm is available for clustering, compression, classification and related optimization problems. DA has the features of avoiding many poor local optimum and being appropriate for different structures. The columns of 3 to 6 show the best published results of J-Means, VNS+, GAUHQ, and SS methods on the two benchmark data sets u1060 and pcb3038. The column "DA" and "EXAMCE" show the results reported in the Christou's work.
The experimental results of the SCE algorithm are presented in the final column. For the data set u1060, our method can obtain the best results except that the value of $k$ is 30. Compare with the EXAMCE method, our results gives the same or ever better performance for the different number of cluster centers. For the pcb3038 data set, the SCE algorithm shows good performance in that 4 out of 5 total cases and only in the case of $k=50$, it finds a worse result than the EXAMCE algorithm. The VNS+ algorithm gives high quality results on the u1060 data set. GAUHQ and SS can work well for some cases. The J-Means and DA algorithm do not perform as well as the other algorithms. From an overall perspective, our method can obtain the best performance on these two data sets.
\begin{table*}[!t]
\begin{center}
\renewcommand{\arraystretch}{1}
\caption{The MSSC values of different algorithms on Wine, SIFT and USPS\label{table:4}}
\centering
\begin{tabular}{c|c|c|c|c|c|c|c|c|c}
\hline

\multirow{2}{*}{k}
   & \multicolumn{9}{c}{Wine}\\
   \cline{2-10}
   & K-means & K-means++ & CSPA & HGPA & MCLA & EAC & EBA & EXAMCE & SCE\\
\hline
5  & 9.1638E+05 & 9.8825E+05 & 1.4944E+06 & 1.3094E+06 & 1.0267E+06 & 9.1638E+05 & 9.1638E+05 & \textbf{9.1637E+05} & \textbf{9.1637E+05} \\
10 & 2.3109E+05 & 2.4327E+05 & 7.2752E+05 & 5.0143E+05 & 4.3318E+05 & 2.3570E+05 & 2.6295E+05 & 2.4327E+05 & \textbf{2.1788E+05} \\
15 & 1.0929E+05 & 1.1536E+05 & 3.0488E+05 & 2.3044E+05 & 1.4696E+05 & 1.3626E+05 & 3.3197E+05 & 1.0892E+05 & \textbf{1.0866E+05} \\
20 & 7.5571E+04 & 7.4960E+04 & 2.0859E+05 & 1.2110E+05 & 1.0925E+05 & 1.0935E+05 & 3.3081E+05 & 6.8077E+04 & \textbf{6.7746E+04}  \\
\hline

\multirow{2}{*}{k}
   & \multicolumn{9}{c}{SIFT}\\
   \cline{2-10}
   & K-means & K-means++ & CSPA & HGPA & MCLA & EAC & EBA & EXAMCE & SCE\\
\hline
   5  & 3.7599E+07 & 3.7603E+07 & 3.8373E+07 & 4.0835E+07 & 3.7611E+07 & 4.0588E+07 & 3.7718E+07 & \textbf{3.7595E+07} & 3.7596E+07\\
   10 & 3.4178E+07 & 3.4232E+07 & 3.4846E+07 & 3.5211E+07 & 3.4426E+07 & 4.0112E+07 & 3.6634E+07 & 3.4008E+07 & \textbf{3.3997E+07}\\
   15 & 3.2468E+07 & 3.2425E+07 & 3.2664E+07 & 3.3613E+07 & 3.2500E+07 & 3.9677E+07 & 3.6829E+07 & 3.1869E+07 & \textbf{3.1866E+07}\\
   20 & 3.1078E+07 & 3.1044E+07 & 3.1484E+07 & 3.2522E+07 & 3.1031E+07 & 3.8794E+07 & 3.5118E+07 & 3.0292E+07 & \textbf{3.0234E+07}\\
   30 & 2.8783E+07 & 2.8941E+07 & 2.9253E+07 & 3.0087E+07 & 2.8937E+07 & 3.9244E+07 & 3.4999E+07 & 2.7811E+07 & \textbf{2.7764E+07}\\
   40 & 2.7296E+07 & 2.7050E+07 & 2.7712E+07 & 2.8513E+07 & 2.7205E+07 & 3.7494E+07 & 3.2558E+07 & \textbf{2.5845E+07} & 2.5852E+07\\
\hline

\multirow{2}{*}{k}
   & \multicolumn{9}{c}{USPS}\\
   \cline{2-10}
   & K-means & K-means++ & CSPA & HGPA & MCLA & EAC & EBA & EXAMCE & SCE\\
\hline
5 & 4.3605E+04 & 4.3599E+04 & 4.5981E+04 & 6.2585E+04 & 4.5265E+04 & 4.4622E+04 & 4.5830E+04 & 4.2263E+04 & \textbf{4.2262E+04}\\
6 & 4.8640E+04 & 4.8641E+04 & 4.9065E+04 & 4.8308E+04 & 4.7636E+04 & 5.0901E+04 & 4.8792E+04 & \textbf{4.7475E+04} & \textbf{4.7475E+04} \\
7 & 5.2182E+04 & 5.2198E+04 & 5.4404E+04 & 5.8245E+04 & 5.2857E+04 & 6.0906E+04 & 5.3378E+04 & 5.2178E+04 & \textbf{5.2168E+04}\\
8 & 5.8277E+04 & 5.8278E+04 & 6.0505E+04 & 9.5456E+04 & 5.8345E+04 & 6.0083E+04 & 5.8610E+04 & \textbf{5.8276E+04} & \textbf{5.8276E+04} \\
9 & 7.2890E+04 & 7.2893E+04 & 7.4367E+04 & 1.1817E+05 & 7.3079E+04 & 7.5225E+04 & 7.4663E+04 & 7.2886E+04 & \textbf{7.2880E+04} \\
10& 7.4083E+04 & 7.4099E+04 & 7.6806E+04 & 1.3010E+05 & 7.4367E+04 & 1.0796E+05 & 7.5513E+04 & 7.4080E+04 & \textbf{7.4070E+04}\\
\hline

\end{tabular}
\end{center}
\end{table*}

\begin{table*}[!t]
\begin{center}
\renewcommand{\arraystretch}{1}
\caption{The MSSC values of different algorithms on FEI, Yale, PIE and OLIVETTIFACES\label{table:5}}
\centering
\begin{tabular}{c|c|c|c|c|c|c|c|c|c}
\hline
\multirow{2}{*}{k}
   & \multicolumn{9}{c}{FEI}\\
   \cline{2-10}
   & K-means & K-means++ & CSPA & HGPA & MCLA & EAC & EBA & EXAMCE & SCE\\
\hline
   5  & \textbf{2.4428E+07} & 2.5883E+07 & 3.6297E+07 & 3.6006E+07 & \textbf{2.4428E+07}	& 2.4535E+07 & 2.6078E+07 & 2.4535E+07 & \textbf{2.4428E+07}\\
   10 & 4.2540E+07 & 4.3383E+07 & 5.7561E+07 & 4.9932E+07 & 4.2897E+07 & 4.4702E+07 & 4.8545E+07 & 4.1805E+07 & \textbf{4.1588E+07}  \\
   15 & 6.2283E+07 & 6.4173E+07 & 7.9166E+07 & 6.5558E+07 & 6.3923E+07 & 6.7050E+07 & 6.9582E+07 & 6.1256E+07 & \textbf{6.0896E+07}  \\
   20 & 7.4932E+07 & 7.9927E+07 & 9.2355E+07 & 8.2366E+07 & 7.6300E+07 & 8.2258E+07 & 9.3183E+07 & 7.4380E+07 & \textbf{7.3490E+07}  \\
   30 & 1.2216E+08 & 1.2356E+08 & 1.4701E+08 & 1.3215E+08 & 1.2688E+08 & 1.3665E+08 & 1.4321E+08 & 1.1782E+08 & \textbf{1.1648E+08}  \\
   40 & 1.6694E+08 & 1.7145E+08 & 1.9657E+08 & 1.8421E+08 & 1.7007E+08 & 2.0151E+08 & 2.2494E+08 & 1.6314E+08 & \textbf{1.6117E+08} \\
   50 & 2.0210E+08 & 2.0558E+08 & 2.1930E+08 & 2.2247E+08 & 2.0474E+08 & 2.3944E+08 & 2.9895E+08 & 1.9439E+08 & \textbf{1.9361E+08}  \\
\hline

\multirow{2}{*}{k}
   & \multicolumn{9}{c}{Yale}\\
   \cline{2-10}
   & K-means & K-means++ & CSPA & HGPA & MCLA & EAC & EBA & EXAMCE & SCE\\
\hline
5  & 7.2558E+07 & 7.3188E+07 & 8.0699E+07 & 7.6935E+07 & 7.3609E+07 & 7.2549E+07 & 7.4138E+07 & 7.2409E+07 & \textbf{7.2197E+07} \\
10 & 1.4489E+08 & 1.4521E+08 & 1.5189E+08 & 1.4713E+08 & 1.4460E+08 & 1.4918E+08 & 1.5141E+08 & \textbf{1.4154E+08} & \textbf{1.4154E+08}\\
15 & 2.1099E+08 & 2.1220E+08 & 2.2055E+08 & 2.1904E+08 & 2.1230E+08 & 2.3588E+08 & 2.2959E+08 & 2.0556E+08 & \textbf{2.0490E+08}\\
\hline

\multirow{2}{*}{k}
   & \multicolumn{9}{c}{PIE}\\
   \cline{2-10}
   & K-means & K-means++ & CSPA & HGPA & MCLA & EAC & EBA & EXAMCE & SCE\\
\hline
    5 & 1.3283E+09 & \textbf{1.3145E+09} & 1.3566E+09 & 1.3381E+09 & 1.3322E+09 & 1.3327E+09 & 1.3228E+09 & 1.3356E+09 & 1.3356E+09 \\
   10 & 2.1158E+09 & 2.1459E+09 & 2.1885E+09 & 2.1596E+09 & 2.1477E+09 & 2.1960E+09 & 2.1400E+09 & 2.1160E+09 & \textbf{2.1155E+09} \\
   15 & 3.1964E+09 & 3.3608E+09 & 3.2266E+09 & 3.1229E+09 & \textbf{3.1019E+09} & 3.3575E+09 & 3.0338E+09 & 3.3386E+09 & 3.2079E+09 \\
   20 & 4.6002E+09 & 4.6585E+09 & 4.5647E+09 & 4.6337E+09 & 4.5971E+09 & 4.6033E+09 & 4.5752E+09 & 4.5878E+09 & \textbf{4.5206E+09} \\
   30 & 6.7074E+09 & 6.8945E+09 & 6.8552E+09 & 6.7073E+09 & 6.6011E+09 & 6.4689E+09 & \textbf{6.4128E+09} & 6.9116E+09 & 6.6145E+09 \\
   40 & 8.6259E+09 & 8.3230E+09 & 8.4262E+09 & 8.3916E+09 & 8.5360E+09 & \textbf{8.1368E+09} & 8.1816E+09 & 8.5113E+09 & 8.3803E+09 \\
   50 & 1.0453E+10 & 1.0456E+10 & 1.0895E+10 & 1.0704E+10 & 1.0539E+10 & 1.0799E+10 & 1.0335E+10 & 1.0630E+10 & \textbf{1.0229E+10} \\
\hline

\multirow{2}{*}{k}
   & \multicolumn{9}{c}{OLIVETTIFACES}\\
   \cline{2-10}
   & K-means & K-means++ & CSPA & HGPA & MCLA & EAC & EBA & EXAMCE & SCE\\
\hline
5  & 1.0450E+08 & 1.0404E+08 & 1.0441E+08 & 1.0703E+08 & 1.0420E+08 & 1.0594E+08 & 1.1074E+08 & 1.0358E+08 & \textbf{1.0325E+08} \\
10 & 2.1560E+08 & 2.0944E+08 & 2.1467E+08 & 2.1201E+08 & 2.1048E+08 & 2.3362E+08 & 2.3560E+08 & 2.0725E+08 & \textbf{2.0704E+08} \\
15 & 2.9100E+08 & 2.8825E+08 & 2.9593E+08 & 2.9478E+08 & 2.8773E+08 & 3.1234E+08 & 3.2069E+08 & 2.8107E+08 & \textbf{2.8030E+08} \\
20 & 3.9838E+08 & 3.9332E+08 & 4.0497E+08 & 4.0227E+08 & 3.8721E+08 & 4.4675E+08 & 4.5583E+08 & 3.7733E+08 & \textbf{3.7533E+08} \\
30 & 5.4822E+08 & 5.4642E+08 & 5.5499E+08 & 5.6266E+08 & 5.2877E+08 & 6.2533E+08 & 6.4782E+08 & 5.1712E+08 & \textbf{5.1627E+08} \\
40 & 7.3258E+08 & 7.1531E+08 & 7.5848E+08 & 7.4026E+08 & 7.0749E+08 & 7.5988E+08 & 9.0079E+08 & \textbf{6.8572E+08} & 6.8809E+08 \\
\hline

\end{tabular}
\end{center}
\end{table*}

Furthermore, to verify the effectiveness of our method in capturing the manifold structure of the raw data, we test our proposed SCE on the other seven datasets, and compare it with K-means, K-means++ \cite{arthur2007k}, CSPA, HGPA, MCLA, Evidence Accumulation (EAC) \cite{fred2005combining}, Embedding-based Approach (EBA) \cite{franek2014ensemble} and EXAMCE. The K-means and K-means++ are the single-run algorithms, and the others are cluster ensemble methods. The K-means++ algorithm chooses the initial cluster centers using a probabilistic model for keeping the centers with large distance and then the K-means algorithm is applied starting with the initial centers to find the final solution. CSPA, HGPA and MCLA have been introduced in related work. EAC is a kind of co-association method. The EBA method transfers the initial clustering results into vector space and uses the concept of means of clustering to produce the consensus clustering result. For the K-means and K-means++ algorithm, we run 100 times for each method and choose the best result as their MSSC values. Table \ref{table:4} and \ref{table:5} give the experimental results on SIFT, Wine, handwritten digit database and face data sets.
\begin{figure*}[t]
\centerline{\includegraphics[width=5.0in]{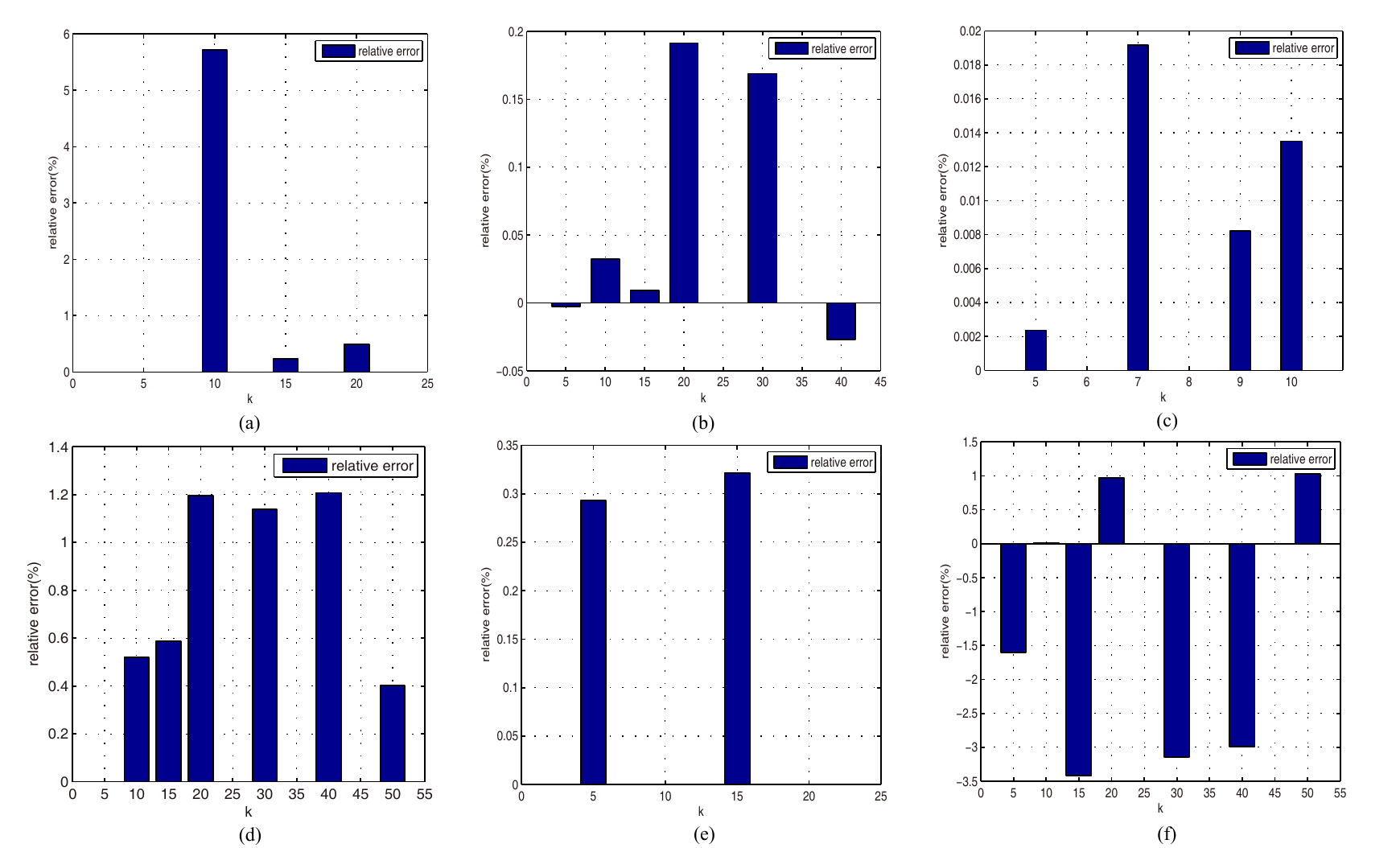}}
\caption{The relative error between our method and the second best solution. The first row (a)-(c) represent the results on Wine, SIFT and USPS data sets. The second row (d)-(f) give the relative error on FEI, Yale and PIE data sets.}
\label{li8}
\end{figure*}
\begin{figure}[t]
\centerline{\includegraphics[width=1.7in]{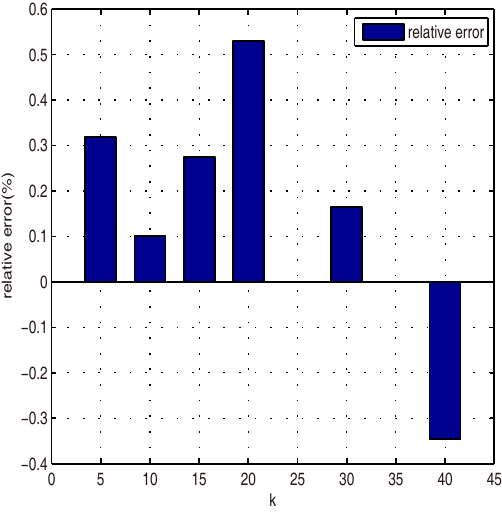}}
\caption{The relative error between our method and the second best solution on Olivettifaces.}
\label{li9}
\end{figure}

From Table \ref{table:4}, we can see that our method can obtain the best clustering results in all situations for Wine and USPS datasets. For the SIFT dataset, our method obtains a better result in 4 out of 6 total cases. As shown in Table \ref{table:5}, our algorithm gives strictly better performance for FEI than the other methods in 6 out of 7 total cases and has the same MSSC value with the K-means algorithm and MCLA when the value of $k$ equals to 5. For the Yale dataset, the SCE algorithm can still achieve the best clustering result with different values of $k$.
For PIE, we randomly sample a subset from original images and conduct clustering algorithm on this subset. After obtaining cluster centers, we cluster all images in PIE data set into $k$ clusters by finding the nearest neighbor for every image and calculate the corresponding MSSC value. Although other clustering algorithms obtain the best solution for some cases, our method gives better quality solutions through balancing all solutions which means the SCE algorithm can still capture the manifold structure for big data set.
For the Olivettifaces dataset, our method outperform the other methods for all cases except for the situation of $k = 40$ where the MSSC value of our method is a bit larger than the result of EXAMCE.

Although we run the K-means and K-means++ algorithm for many times and take the best solution as their MSSC values, K-means and K-means++ method give the poor clustering result for almost all the data sets. For other ensemble cluster methods, they can give a better performance for some cases. From the results in Table \ref{table:4} and \ref{table:5}, we can see that our proposed SCE is effective to capture the manifold structure of the data.

\begin{table*}[!t]
\begin{center}
\renewcommand{\arraystretch}{1}
\caption{The AC (\%) of different algorithms on Yale and USPS data sets\label{table:AC}}
\centering
\begin{tabular}{c|c|c|c|c|c|c|c|c|c}
\hline
\multirow{2}{*}{k}
   & \multicolumn{9}{c}{Yale}\\
   \cline{2-10}
   & K-means & K-means++ & CSPA & HGPA & MCLA & EAC & EBA & EXAMCE & SCE\\
\hline
5	& 67.27 & 63.64 & 69.09 & \textbf{70.91} & 67.27 & 65.45 & 67.27 & 65.45 & 69.09\\
10	& 53.64 & 46.36 & 50.91 & 50.91	& 53.64 & 43.64 & 52.73 & 50.00 & \textbf{56.36}\\
15	& 50.30 & 42.42	& 50.30 & 48.49	& 44.85 & 33.94 & 44.24 & 46.06 & \textbf{50.90}\\

\hline

\multirow{2}{*}{k}
   & \multicolumn{9}{c}{USPS}\\
   \cline{2-10}
   & K-means & K-means++ & CSPA & HGPA & MCLA & EAC & EBA & EXAMCE & SCE\\
\hline
5  & 91.20 & 91.20 & 91.20 & 20.00 & 91.20 & \textbf{91.40 }& 91.30 & 91.20 & \textbf{91.40}\\
6  & 87.08 & 86.75 & \textbf{87.67} & 61.25 & 86.75 & 47.33 & 86.92 & 86.83 & 87.42\\
7  & 85.29 & 84.07 & 83.93 & 66.64 & 80.79 & 39.93 & 84.86 & 85.29 & \textbf{85.43}\\
8  & 86.06 & 85.19 & 84.50 & 12.50 & 86.19 & 25.25 & 81.00 & 86.19 & \textbf{86.25}\\
9  & 82.78 & \textbf{83.72} & 80.56 & 53.56 & 82.11 & 22.50 & 75.78 & 83.50 & 83.61\\
10 & 76.55 & 75.95 & 76.05 & 10.10 & 72.40 & 18.40 & 76.55 & 77.00 & \textbf{77.05}\\

\hline

\end{tabular}
\end{center}
\end{table*}

\begin{figure}[t]
\centerline{\includegraphics[width=2.4in]{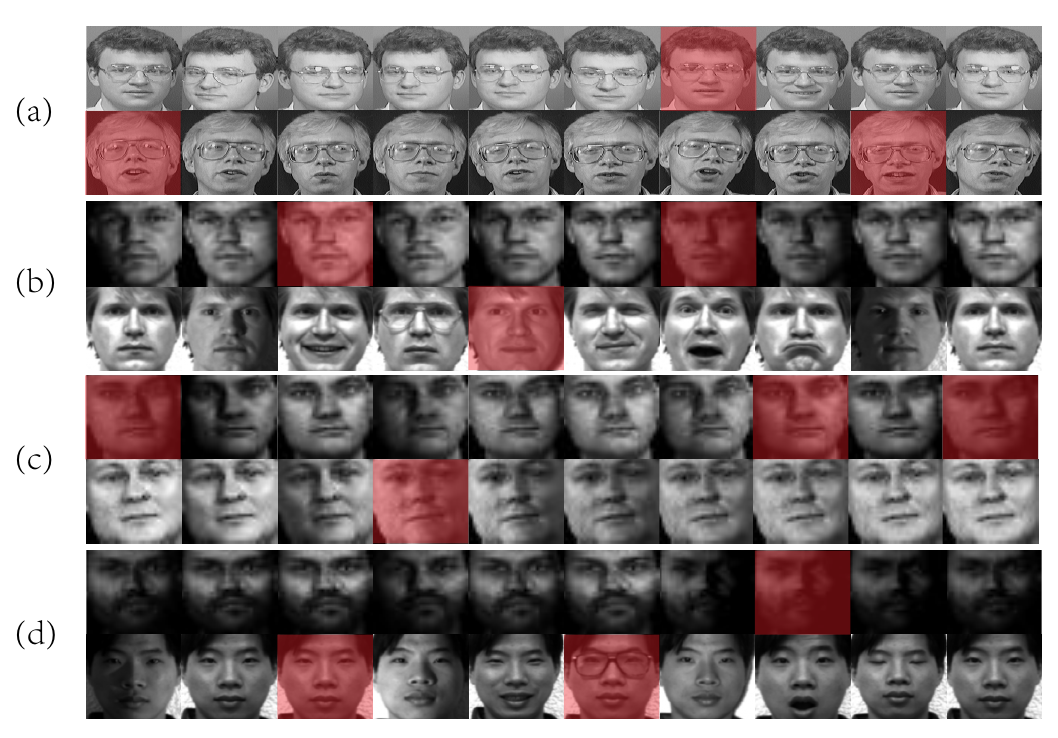}}
\caption{Partial visualization results on face databases. $(a)\sim(d)$ represent experimental results on Olivettifaces, Yale, FEI and PIE data set, respectively.}
\label{li10}
\end{figure}

Fig. \ref{li8} and Fig. \ref{li9} show the relative error (RE) of our method with the best result of other algorithms based on the value in Table \ref{table:4} and \ref{table:5}. Let $R_1$ represents the MSSC value of our method and $R_2$ represents the best value of other algorithms. Then the formula of RE (\%) is given by $RE = (R_2-R_1)/R_2\times100$.
Since the lower MSSC value means better solution, we use the formula of ${R_2} - {R_1}$ in numerator for showing the improvement with positive value.
From Fig. \ref{li8} and Fig. \ref{li9}, we can see the similar results in a more intuitive style. Horizontal ordinate represents the different numbers of clusters and vertical coordinate denotes the relative error. Positive values represent our method outperforms the other methods. It is clear to find that our method obtains in 27 cases better quality results comparing to other methods and is slightly worse only 7 times. For the Wine data set, our SCE algorithm is more than 1 percent better than the second best MSSC value for $k=10$. Our method obtains much higher value on the FEI data set.
Fig. \ref{li10} shows partial visualization results on face databases. The image with red tag means that it is assigned to a wrong cluster rather than the cluster to which it belongs.

Finally, we use clustering accuracy (AC) \cite{7297854} to further evaluate the effectiveness of our method on the labeled data sets. Due to space limitation, we choose Yale and USPS as the experimental datasets. Table \ref{table:AC} gives the corresponding clustering results. Note that our method gives the best performance in 6 out of 9 total cases. For other three cases, our method obtains the second best results which are a slightly lower than the best results. So the experimental results show the effectiveness of our method for the labeled data sets.

\subsection{Computational Complexity}
The time cost of each cluster ensemble method consists of those of computing the initial clustering results and combining them. In the first stage, we use MMDA to produce the initial clustering results. The time complexity of MMDA is $O(nkr)$, where $n$ is the number of data points, $k$ is the number of cluster centers and $r$ is the number of initial clustering results. The second stage is to combine the initial clustering results and further produce the final result. The mainly time cost of the second stage is to solve the equation (\ref{eq:5}). During solving the problem (\ref{eq:5}), we adopt the interior point convex algorithm which is an iterative method with time cost $O(kr{t_1})$, where ${t_1}$ is the number of iterations. Because Algorithm \ref{alg:1} needs to reconstruct the objective function and repeat the step of solving the problem (\ref{eq:5}), the time complexity of the second stage is $O(kr{t_1}{t_2})$, where $t_2$ is the iterations of solving the problem (\ref{eq:5}). Therefore, the time complexity of our method is $O(nkr + kr{t_1}{t_2})$.

Table \ref{table:Time cost} reports the time complexity of our SCE method and other representative ensemble clustering methods.
The time complexity of initial clustering in the other six cluster ensemble methods is $O(nktr)$ due to that all of them adopt K-means or K-means++ to produce the initial clustering results. However, our SCE method uses MMDA in the initial clustering step. So our method has a lower time cost in producing the initial results. Besides, it can be seen that CSPA, HGPA, MCLA and EBA are slower and can be impractical for a lager $n$. For EAC, it is apparently slower than our SCE method when $max\{t_1,t_2\}<min\{k,r\}$. For  EXAMCE, $t_3$ is the iterations of iterative optimization algorithm and $t_4$ is the iterations of solving its objective function for finding the final result. So, its time complexity is lower than SCE in the second stage. However, the whole time cost of EXAMCE and SCE depends on the specific values of $t_1\sim t_4$.
\begin{table}[!t]
\begin{center}
\renewcommand{\arraystretch}{1}
\caption{The time complexity of our SCE method and other representative ensemble clustering methods}\label{table:Time cost}
\centering
\begin{tabular}{*{1}{c|} c}
\hline
Method & Time Complexity\\
\hline\hline
CSPA & $O(nktr + {n^2}kr)$\\
\hline
HGPA & $O(nktr + nkr)$\\
\hline
MCLA & $O(nktr + n{k^2}{r^2})$\\
\hline
EAC & $O(nktr + {k^2}{r^2})$\\
\hline
EBA & $O(nktr + {n^2}kr)$\\
\hline
EXAMCE & $O(nktr + {t_3}{t_4})$\\
\hline
SCE & $O(nkr + kr{t_1}{t_2})$\\
\hline
\end{tabular}
\end{center}
\end{table}

\section{Conclusion}
By investigating cluster ensemble, we in this paper propose a Structural Cluster Ensemble (SCE) algorithm for data partitioning formulated as a set-covering problem. Firstly, the Max-Min Distance Algorithm (MMDA) is employed to obtain different initial clustering results which has a low time complexity. So our SCE method has a lower time cost in producing the initial results. Then, we construct a Laplacian regularized objective function for capturing the structure information among clusters. So the close clusters on a manifold can keep the same values in the solution. Furthermore, we incorporate the discriminative information into our proposed objective function. It can guarantee the clusters we chosen have high dispersion degree which follows the cluster criterion. Therefore, our proposed method is capable of obtaining a high quality fusion clustering result. Finally, we compare our method with thirteen representative methods. Experimental results on ten popular data sets reveal that our proposed SEC algorithm is effective and robust, especially for image datasets.

\ifCLASSOPTIONcaptionsoff
  \newpage
\fi

{
\bibliographystyle{IEEEtran}
\bibliography{egbib}
}

\begin{IEEEbiographynophoto}{Xuelong Li}
(M'02-SM'07-F'12) is a Full Professor with the Center for
Optical Imagery Analysis and Learning, State Key Laboratory of Transient
Optics and Photonics, Xi¡¯an Institute of Optics and Precision Mechanics,
Chinese Academy of Sciences, Xi¡¯an, China.
\end{IEEEbiographynophoto}
\ \\
\ \\
\ \\

\begin{IEEEbiography}[{\includegraphics[width=1in,height=1.32in,clip,keepaspectratio]{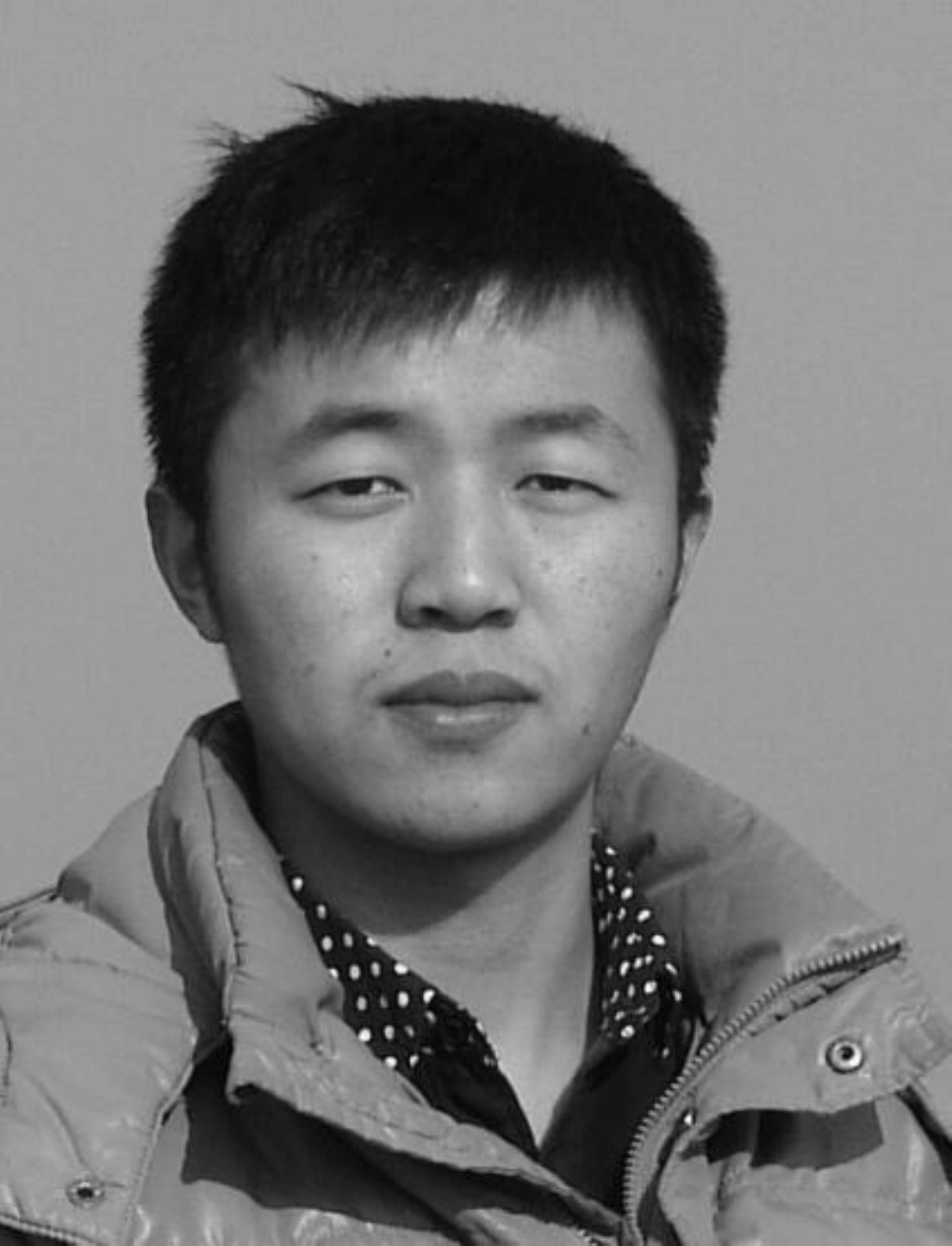}}]{Quanmao Lu}
is currently working toward the Ph.D. degree in the Center for Optical Imagery Analysis and Learning, State Key Laboratory of Transient Optics and Photonics, Xi'an Institute of Optics and Precision Mechanics, Chinese Academy of Sciences, Xi'an, China.
His current research interests include machine learning and computer vision.
\end{IEEEbiography}

\ \\
\ \\
\ \\
\ \\
\ \\
\ \\

\begin{IEEEbiography}[{\includegraphics[width=1in,height=1.32in,clip,keepaspectratio]{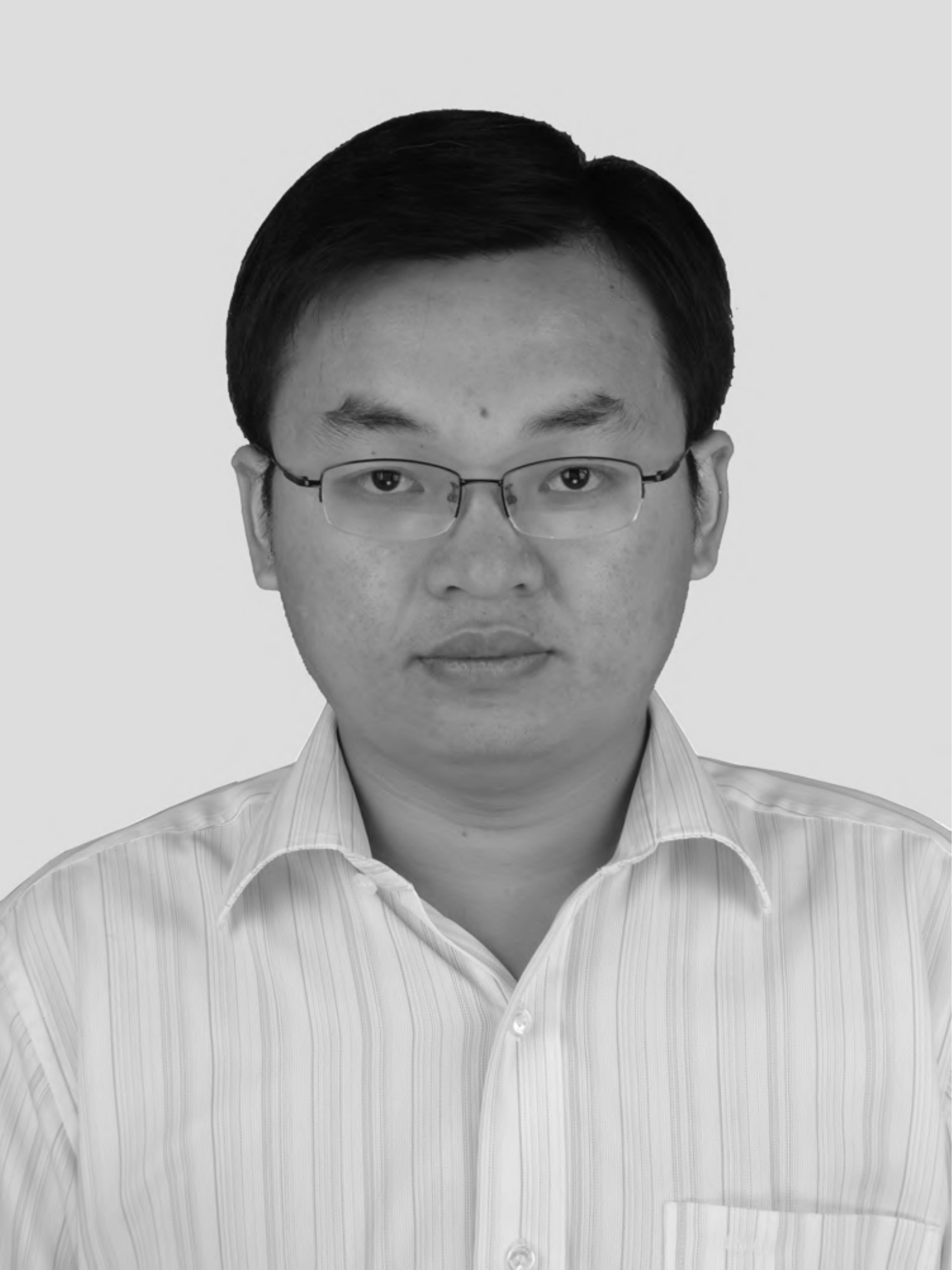}}]{Yongsheng Dong}%
(M'14) received his Ph. D. degree in applied mathematics from Peking University in 2012. He was a postdoctoral fellow with the Center for Optical Imagery Analysis and Learning, State Key Laboratory of Transient Optics and Photonics, Xi'an Institute of Optics and Precision Mechanics, Chinese Academy of Sciences, Xi'an, China from 2013 to 2016. He is currently an associate professor with the Information Engineering College, Henan University of Science and Technology. His current research interests include pattern recognition, machine learning, and computer vision.

He has authored and co-authored over 20 journal and conference papers, including IEEE TIP, IEEE TCYB, IEEE SPL and ACM TIST. Meanwhile, he has served as a reviewer for over 30 international prestigious journals and conferences, such as IEEE TNNLS, IEEE TIP, IEEE TCYB, IEEE TIE, IEEE TSP, IEEE TKDE, IEEE TCDS, ACM TIST, BMVC and ICIP. He has served as a Program Committee Member for over ten international conferences. He is a member of the IEEE, ACM and CCF.

\end{IEEEbiography}

\begin{IEEEbiography}[{\includegraphics[width=1in,height=1.32in,clip,keepaspectratio]{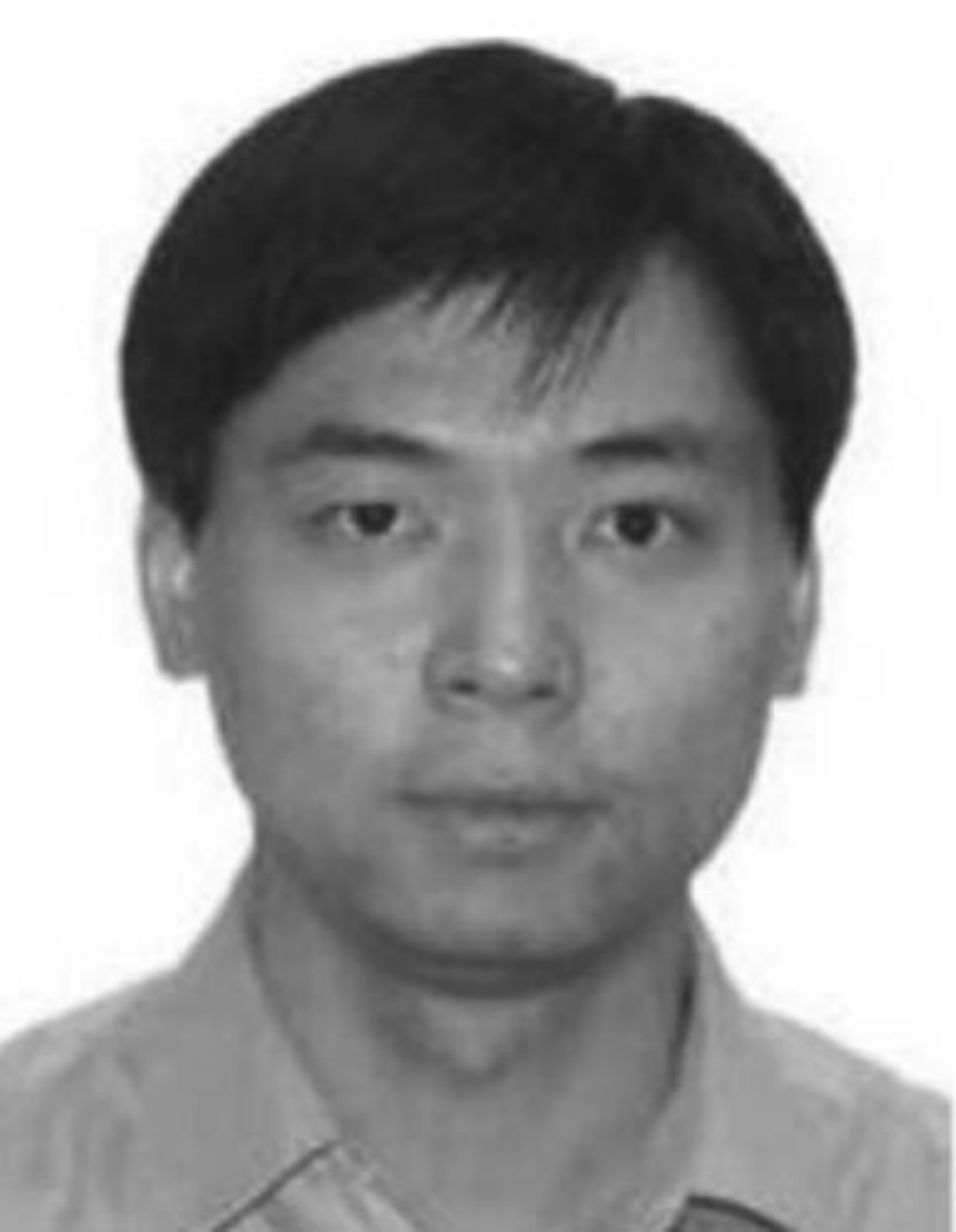}}]{Dacheng Tao}%
(F'15) is Professor of Computer Science and ARC Future Fellow in the School of Information Technologies and the Faculty of Engineering and Information Technologies, and the Founding Director of the UBTech Sydney Artificial Intelligence Institute at the University of Sydney. He was Professor of Computer Science and Director of Centre for Artificial Intelligence in the University of Technology Sydney. He mainly applies statistics and mathematics to Artificial Intelligence and Data Science. His research interests spread across computer vision, data science, image processing, machine learning, and video surveillance. His research results have expounded in one monograph and 500+ publications at prestigious journals and prominent conferences, such as IEEE T-PAMI, T-NNLS, T-IP, JMLR, IJCV, NIPS, CIKM, ICML, CVPR, ICCV, ECCV, AISTATS, ICDM; and ACM SIGKDD, with several best paper awards, such as the best theory/algorithm paper runner up award in IEEE ICDM¡¯07, the best student paper award in IEEE ICDM¡¯13, and the 2014 ICDM 10-year highest-impact paper award. He received the 2015 Australian Scopus-Eureka Prize, the 2015 ACS Gold Disruptor Award and the 2015 UTS Vice-Chancellor¡¯s Medal for Exceptional Research. He is a Fellow of the IEEE, OSA, IAPR and SPIE.

\end{IEEEbiography}

\ \\
\ \\
\ \\
\ \\
\ \\
\ \\
\ \\
\ \\
\ \\
\ \\
\ \\
\ \\
\ \\
\ \\
\ \\
\ \\
\ \\
\ \\

\end{document}